\documentclass[preprint,12pt]{elsarticle}

\usepackage{amsmath,amssymb}
\usepackage[utf8x]{inputenc}
\usepackage{textcomp,marvosym}
\usepackage[table]{xcolor}

\usepackage{array}

\usepackage{amsfonts}
\usepackage{algorithmic}
\usepackage{graphicx}
\usepackage{multirow}
\usepackage{comment}
\usepackage{csquotes}
\usepackage[]{algorithm2e}
\usepackage{float}
\usepackage{blindtext}
\usepackage{ulem}
\usepackage{colortbl}
\usepackage{hhline}

\usepackage{nameref}

\usepackage{url}

\usepackage{breakurl}
\usepackage{breakcites}

\usepackage[font=small,skip=0pt]{caption}
\usepackage{setspace}

\journal{arXiv.org}

\begin{document}

\begin{frontmatter}



\title{Classification with 2-D Convolutional Neural Networks for breast cancer diagnosis}


\author{Anuraganand Sharma}
\address{School of Computing Information \& Mathematical Sciences \\
              The University of the South Pacific\\
              Suva, Fiji\\
              {sharma\_au@usp.ac.fj} }

\author{Dinesh Kumar}
\address{              Faculty of Science \& Technology\\
              University of Canberra, Canberra, ACT, Australia\\
              {dinesh.kumar@canberra.edu.au}}

\begin{abstract}
 Breast cancer is the most common cancer in women. Classification of cancer/non-cancer patients with clinical records requires high sensitivity and specificity for an acceptable diagnosis test. The state-of-the-art classification model - Convolutional Neural Network (CNN), however, cannot be used with clinical data that are represented in 1-D format. CNN has been designed to work on a set of 2-D matrices whose elements show some correlation with neighboring elements such as in image data. Conversely, the data examples represented as a set of 1-D vectors -- apart from the time series data -- cannot be used with CNN, but with other classification models such as Artificial Neural Networks or RandomForest. We have proposed some novel preprocessing methods of data wrangling that transform a 1-D data vector, to a 2-D graphical image with appropriate correlations among the fields to be processed on CNN. We tested our methods on Wisconsin Original Breast Cancer (WBC) and Wisconsin Diagnostic Breast Cancer (WDBC) datasets. To our knowledge, this work is novel on non-image to image data transformation for the non-time series data. The transformed data processed with CNN using VGGnet-16 shows competitive results for the WBC dataset and outperforms other known methods for the WDBC dataset. 
 \end{abstract}


\end{frontmatter}

\section{Introduction}\label{sec:sec1}
In recent times, there are growing interest in the development of machine learning (ML) models for medical datasets due to the advancements in digital technology and improvements in data collection methods. Increasingly, several ML-based systems have been designed as an early warning or diagnostic tool for chronic illnesses, for example diagnosing depression, diabetes and cancer \cite{sourla2012cardiosmart365}. Breast cancer is arguably one of the deadliest forms of cancer amongst women with millions of reported cases around the world of which many cases become fatal \cite{gao2018sd,tsai2020breast}. Breast cancer is caused by abnormal growth of some of the breast cells in the lining of the milk glands or ducts of the breast (ductal epithelium) \cite{noauthor_breast_nodate,noauthor_breast_nodate-1}. Compared to healthy cells, these cells divide more rapidly and accumulate, forming a lump or mass. At this stage, the cells become malignant and may spread through the breast to lymph nodes or other parts of the body.

\subsection{Problem Statement} \label{sec:sec1:probStatement}

The study of breast cancer has attracted considerable attention in the past decades. Improving data collection and storage technologies has resulted in various types and amounts of data collected on breast cancer from around the world. These include data on Ribonucleic Acid  (RNA) signatures for cell mutations that cause breast cancer \cite{kaur2019comparison,larsen2014microarray}, mammogram images \cite{dembrower2019multi,bowyer1996digital} and data on symptoms and diagnosis \cite{dheeru_uci_2019}. Many traditional Computer-Aided Diagnosis (CADx) systems require hand-crafted feature extraction which is a challenging task \cite{sun_enhancing_2017,firmino_computer-aided_2016}. Even conventional ML techniques require the extraction of an optimal set of features manually prior to model training. An extensive review on various feature selection and extraction techniques can be found in \cite{guyon2003introduction, kumar2014feature}. Some commonly used approaches for ML models are Principal Component Analysis (PCA) \cite{fodor2002survey}, information gain \cite{liu_novel_2019}, GA-based feature selection \cite{babatunde2014genetic}, recursive feature elimination (RFE) \cite{darst2018using}, meta-heuristic methods \cite{sharma2020comprehensive} and rough sets \cite{pawlak_rough_2012}. Feature selection and extraction, therefore, is an important consideration in the pre-processing step before applying any ML algorithm such as decision trees, Bayesian models, Support Vector Machines (SVM) and Artificial Neural Networks (ANN). The behavior of ML algorithms and their prediction accuracy is influenced by the choice of features selected \cite{guyon_introduction_2003,singh_feature_2015}. Many times manual feature extraction or knowledge of domain experts is needed to have a good understanding on the relevance of the attributes \cite{mohamad_feature_2004}.

\subsection{Context and Background}\label{sec:sec1:Context}
To address these issues surrounding the use of conventional ML algorithms has propelled the need for new approaches and methods to automatically extract features from large datasets. As a result, Deep Learning (DL) algorithms such as Convolutional Neural Network (CNN or ConvNet) and Recurrent Neural Networks (RNNs) have emerged in recent times that can accept raw data and are automatically able to discover patterns in them \cite{kumar_deep_2019,cui_multi-scale_2016}.

CNN is one of the most popular algorithms for deep learning which is mostly used for image classification, natural language processing, and time series forecasting. Its ability to extract and recognize the fine features has led to the state-of-the-art performance in various application domains such as computer vision, image recognition, speech recognition, natural and language processing \cite{krizhevsky_imagenet_2012,simonyan_very_2014,volokitin_deep_2017}. CNN is an enhancement of a canonical Neural Networks architecture that is specifically designed for image recognition in \cite{lecun_backpropagation_1989}. Since then many variations have been added to the architecture of CNN to enhance its ability to produce remarkable solutions for deep learning problems such as AlexNet \cite{krizhevsky_imagenet_2012}, VGG Net \cite{simonyan_very_2014} and GoogLeNet \cite{szegedy_going_2015}. CNN eliminates the need for manual feature extraction because the features are learned directly by different convolutional layers \cite{guo_simple_2017,krizhevsky_imagenet_2012}. It does not require a separate feature extraction strategy which requires domain expert and other preprocessing techniques where complete features may still not be extracted \cite{indolia_conceptual_2018}. Despite its huge success with image data, CNN is not designed to handle non-image\footnote{All future referencing of non-image data are in non-time series form unless otherwise specified.} data in non-time series form. Arguably, any problem that can represent the correlation of features of a given data example in a single map, maybe attempted via CNN.

\indent CNNs have proven to work best on data that are in 2-D form, such as images and audio spectrograms \cite{dllectures}. This is attributed to the fact that the convolution technique in CNN requires data examples to have at least two dimensions. Conversely, CNN has been explored on application-specific 1-D data as well. These include gene sequencing data such as DNA sequences being treated as text data (sequence of words) \cite{nguyen2016dna}, and signals and sequences in text mining, word detection and natural language processing (NLP) \cite{delakis2008text,xu2015robust}. More specifically, CNN for Time-Series Classification (TSC) has been recently explored with some new methods such as Multi-Scale CNN (MCNN) \cite{cui_multi-scale_2016} and an ensemble of CNN models with AlexNet on Inception-v4 architecture \cite{szegedy_inception-v4_2017,fawaz_inceptiontime_2019}. These methods have made significant improvement in the accuracy of the classifiers with the state-of-the-art ensemble methods such as Flat-COTE and HIVE-COTE \cite{lines_hive-cote_2016,bagnall_time-series_2015}. Moreover, raw time-series data has also been used into 1-D CNN by calculating the area of the signal for convolution with better time complexity and scalability \cite{brownlee_deep_2018,pydata_1d_nodate}. Nonetheless, much data still exists in a 1-D format such as clinical data of medical records, and therefore, opens challenging research questions on whether they can be effectively trained for classification using CNN. This paper is aimed at filling this gap by proposing a novel non-time series 1-D numerical data to 2-D data transformation methods and processing them with CNN. 

\subsection{Motivation}\label{sec:sec1:Motivation}
\indent The main motivation for this paper is to realize the potential of CNN for non-image clinical data for breast cancer because it eliminates the need for manual feature extraction.  The features are learned directly by CNN whereby it also produces state-of-the-art recognition results \cite{alom_state---art_2019}. The key difference between traditional ML and DL is in how features are extracted. Traditional ML approaches use handcrafted engineering features by applying several feature extraction algorithms and then apply the learning algorithms. On the other hand, in the case of DL, the features are learned automatically and are represented hierarchically at multiple levels. This is the strong point of DL against traditional machine learning approaches \cite{alom_state---art_2019}.

\subsection{Hypothesis}\label{sec:sec1:Hypothesis}
\begin{figure}[tb]
	\begin{center}
		\includegraphics[width=0.75\linewidth]{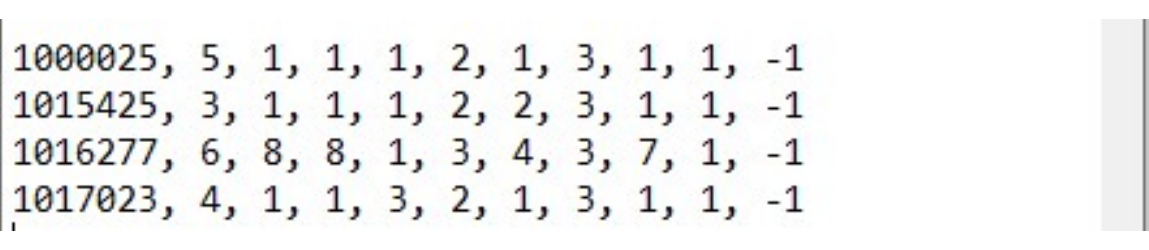}
	\end{center}
	\caption{Snapshot of data file for Breast Cancer dataset WBC from \cite{dheeru_uci_2019}}
	\label{fig:breast_dataset}
\end{figure}

\indent We have proposed some novel methods to transform non-image clinical data of breast cancer to 2-D feature map images in $\mathbb{R}^2$ so that a large set of these kinds of data are not deprived of the services of CNN. This would also encourage other variations and/or methods for text to image transformation to be developed in the future. The scope of this paper is to broaden the usage of CNN to those applications where $d$-dimensional raw data has set of $N$, 1-D data vectors in $\mathbb{R}$ as shown in \figurename\textbf{ }\ref{fig:breast_dataset}. Each row represents a 1-D data vector with $d$ elements where $d, N$ $\geq1$. It is a sample of a Wisconsin Original Breast Cancer dataset (WBC) used in the experiments. This dataset from UCI \cite{dheeru_uci_2019} is a record of medical examination of patients to diagnose breast cancer, where each row is a 1-D vector representing a numerical data example. We demonstrate our method of non-image breast cancer data transformation to image data -- processed in CNN -- produces exceptional results for classification accuracy. 
Some research demonstrates the use of 1-D convolutions on 1D datasets such as data in the form of signals and time sequences \cite{1DECGSignals}. Though this provides a possibility of using 1-D convolutions in this research, our experiments revealed their unsuitability on our experimental datasets. Having applied the data in its raw form into 1-D CNN gave highly unpredictable results.

\indent This paper is organized as follows: Section \ref{sec:sec2} briefly describes the general architecture of CNN. Section \ref{sec:sec3} describes our three proposed methods of data wrangling from non-image Breast Cancer data \cite{dheeru_uci_2019} to image data. Section \ref{sec:sec4} describes the complete methodology of the classification of breast cancer data with CNN. Section {\ref{sec:sec5} shows the experimental results and Section \ref{sec:sec6}} discusses the outcome of the experiments. Lastly, Section \ref{sec:sec7} concludes the paper by summarizing the results and proposing some further extensions to the research.

\section{Convolutional Neural Networks}\label{sec:sec2}
A convolutional neural network (CNN or ConvNet) is a deep learning algorithm designed for computer vision. Its architecture is based on backpropagation artificial neural networks \cite{lecun_backpropagation_1989}. It takes an input image whose each pixel represents input data that goes through a series of the feature selection process through convolution which is later sent to the weighted perceptrons where the learning happens through backpropagation. The major advantage of CNN is its ability to learn the features by itself while in the canonical neural networks feature selection is a separate process where the final accuracy of the model depends on the choice of preprocessing and feature selection methods \cite{khan_guide_2018,saha_comprehensive_2018}. CNN has become a prominent deep learning model with a plethora of literature available on its structure and functionality, however, a brief description of individual layers of CNN is given below.

\subsection{Feature Selection layer}
This layer is a feature extraction layer for CNN which means any additional domain-specific feature selection preprocessing is not required. This layer can be divided into 3 sublayers:
\subsubsection{Convolutional Layer}
This layer directly accepts raw images as input where a set of small filters is convolved over the image to produces one or more feature maps \cite{son_lam_phung_matlab_2009,stutz_understanding_2014}. Convolution happens through sliding the filter across the image while computing the dot product of elements of the filter and image \cite{lichman_uci_2013}. This process results in the extraction of certain features from the image \cite{noauthor_convolutional_nodate}. 
\subsubsection{Activation Layer}
The results of the convolutional layer are passed through an activation function to produce a bounded output. CNN generally uses the Rectified linear unit (ReLU) that converts negative values to 0. It also trains the network several times faster than its counterparts such as $tanh$ \cite{krizhevsky_imagenet_2012}.
\subsubsection{Pooling Layer}
This layer does the downsampling that also reduces the input size along each dimension \cite{noauthor_convolutional_nodate}. Some common pooling methods are average pooling and max pooling where the received image is partitioned into a set of non-overlapping rectangles. Max-pooling and average pooling get only the maximum value and average value of every sub-region respectively. This process downsamples the image \cite{brownlee_gentle_2019,noauthor_convolutional_nodate-2}.
\subsection{Classification Layer}
\begin{figure}[tb]
	\begin{center}
		\includegraphics[width=0.75\linewidth]{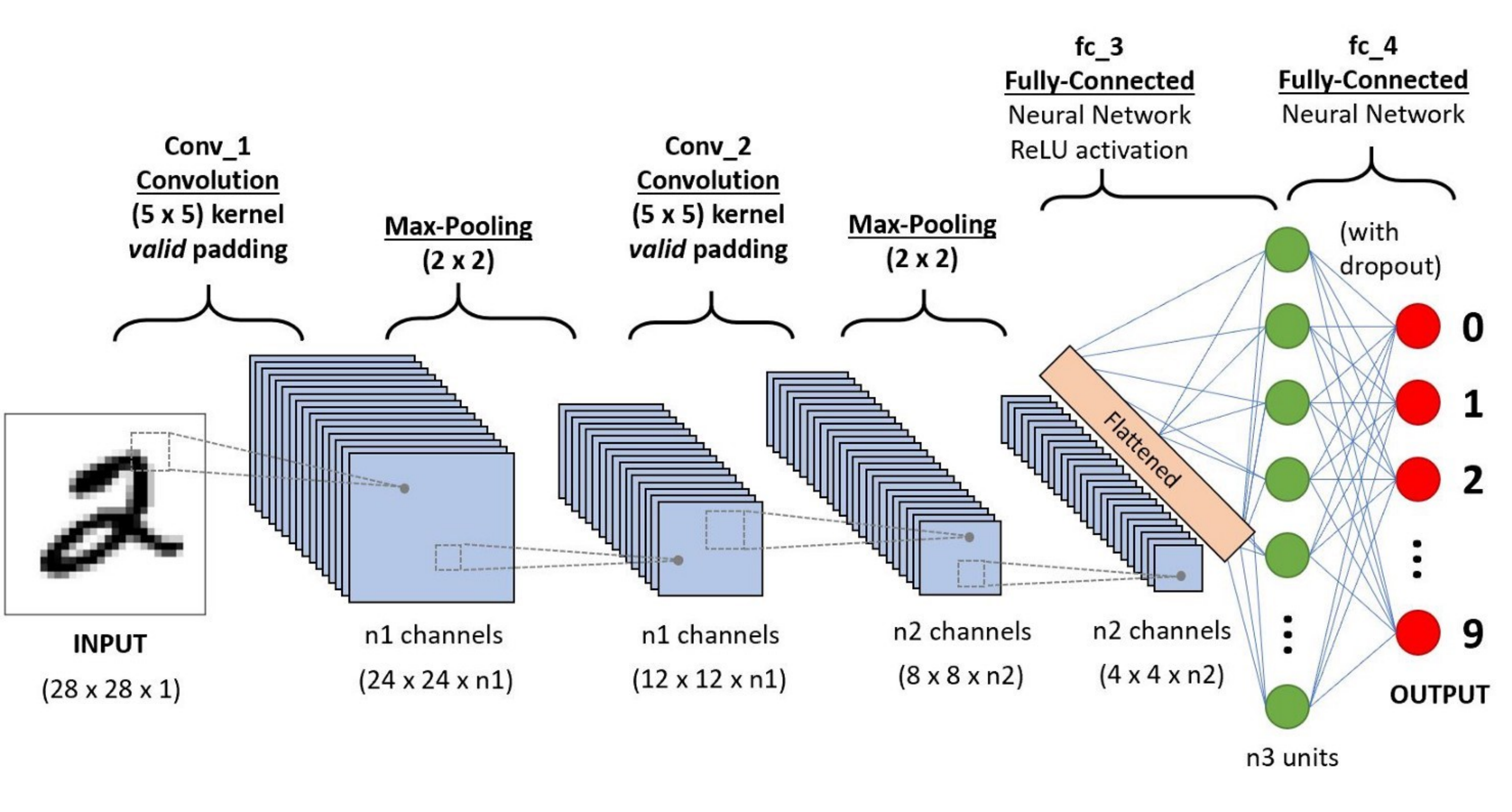}
	\end{center}
	\caption{{\bf A general architecture of CNN} – taken from \cite{saha_comprehensive_2018}.}
	\label{fig:CNN}
\end{figure}

After learning features in the above layer, the architecture of CNN shifts to classification. This fully connected layer is similar to the fully connected network in the conventional neural network models \cite{indolia_conceptual_2018}. The final layer of the CNN architecture uses a classification layer such as softmax to provide the classification output \cite{noauthor_convolutional_nodate}.
The complete architecture of CNN taking an image of number 2 is shown in \figurename\textbf{ }\ref{fig:CNN} (taken from \cite{saha_comprehensive_2018}). The image goes through all the layers which are then classified between values 0 – 9. 

\section{Preprocessing Methods to Transform Numerical Data to Image}\label{sec:sec3}
We have proposed three basic techniques of data wrangling to convert Breast Cancer numerical data to image data. The converted image must reflect some patterns to depict a given class. We have used Wisconsin Original Breast Cancer (WBC) and Wisconsin Diagnostic Breast Cancer (WDBC) datasets from the UCI library \cite{dheeru_uci_2019} for the classification of numerical data in this work.
\subsection{Equidistant Bar Graphs}\label{subsec:equidis}
The bar graph represents the measurement of every feature of a given dataset. There are lots of possibilities of drawing a bar graph but we have used a simplistic approach. The dataset is first normalized to $[0,1]$ then every feature is drawn based on its measured value. The width of the image in pixels is $\psi d+\gamma(d+1)$ where $d$ is total features, $\psi$ is the width of a bar and $\gamma$ is gap between two consecutive bars. The height of the image is normalized to produce a square image. We used $1-$pixel length for $\psi$ and $2-$pixels length for $\gamma$ in our experiments. This produces the square image of size $[3d\times3d]$ approximately. Few data examples of WDBC dataset converted to bar graphs are shown in \figurename\textbf{ }\ref{fig:bargraph} with class labels – Benign and Malignant. The algorithm for this approach is given in the Appendix. 

\begin{figure}[tb]
	\begin{center}
		\includegraphics[width=0.65\linewidth]{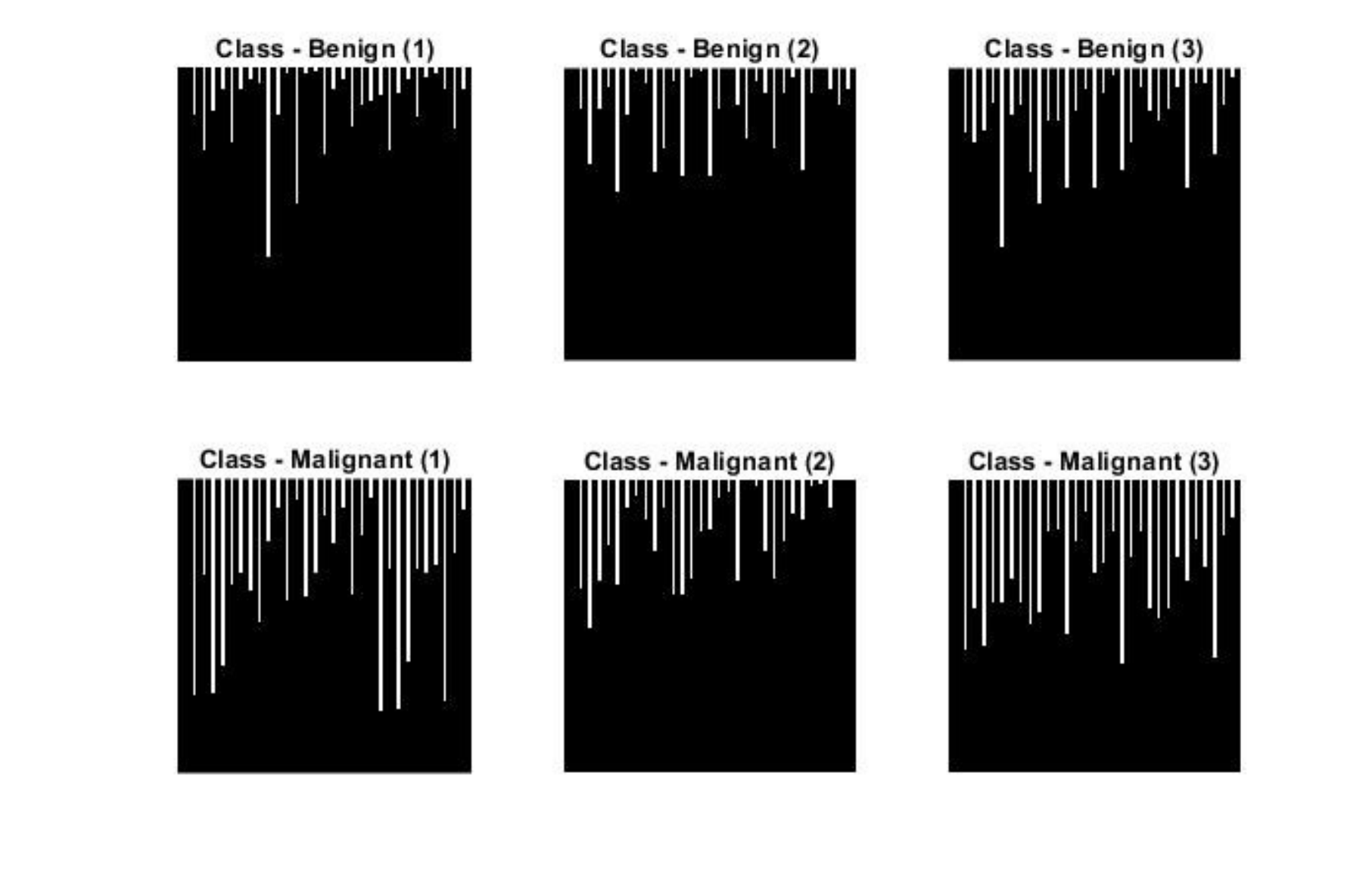}
	\end{center}
	\vspace*{-8mm}
	\caption{Bar graph for some data examples of WDBC dataset.}
	\label{fig:bargraph}
\end{figure}

\begin{figure}[tb]
	\begin{center}
		\includegraphics[width=0.65\linewidth]{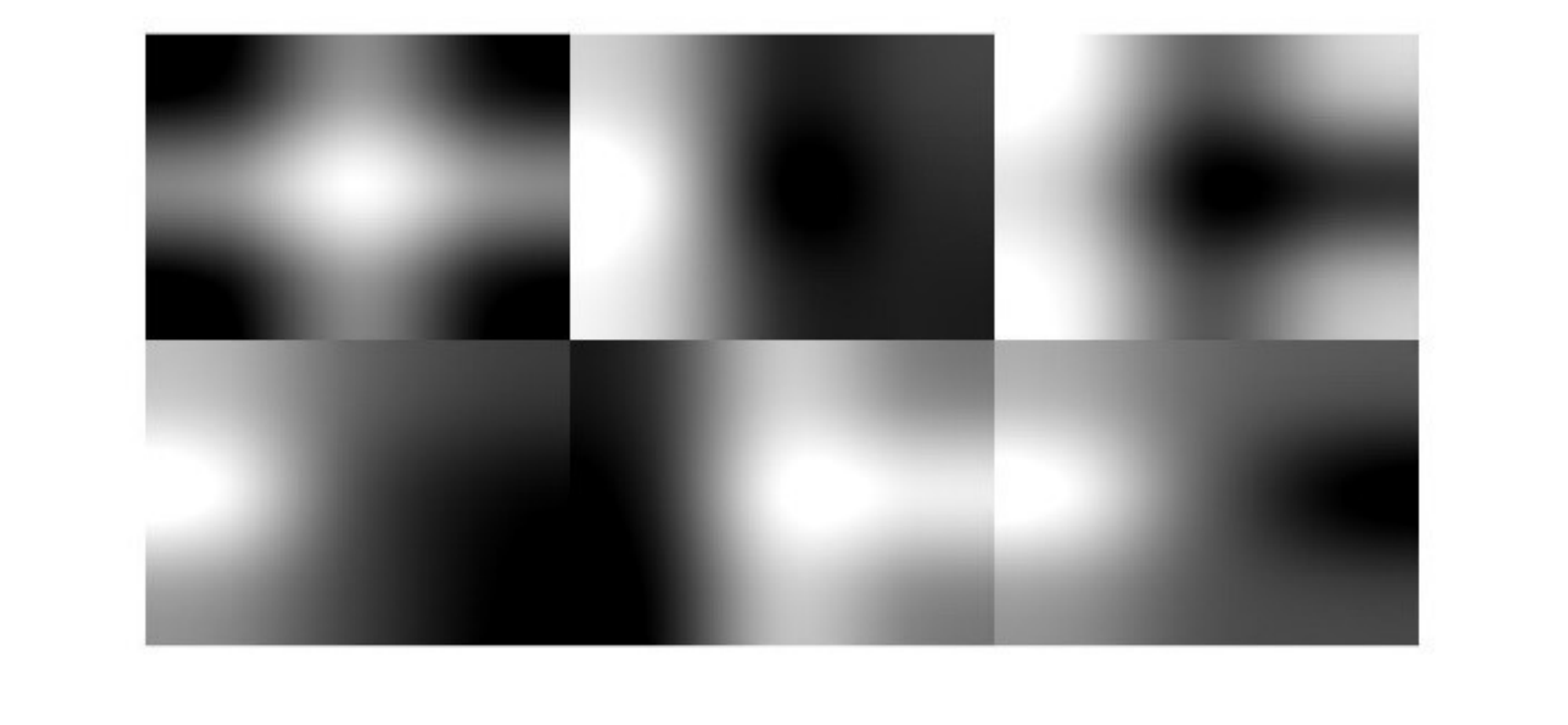}
	\end{center}
	\vspace*{-5mm}
	\caption{Features learned by the first convolutional layer for Breast Cancer dataset.}
	\label{fig:featuresCNN-BreastCancer}
\end{figure}

These pictures are only useful to CNN if they depict a pattern in a convolved image. The first convolutional layer produces 6 features which are shown in \figurename\textbf{ } \ref{fig:featuresCNN-BreastCancer} where some sort of distinguishing features have been reflected.

Intuitively, the \enquote{correct} order of the bars ought to give better results. The datasets of numerical data were reorganized where the related fields were put close to each other according to the order of their similarity. Firstly, a covariance matrix on data fields was generated then each value of the matrix is converted to `rank' that determines how closely one field is related to the other. This is a shortest-path problem where algorithms such as dynamic programming or any metaheuristic algorithm \cite{almufti_historical_2019} such as Genetic Algorithm (GA) \cite{goldberg_genetic_1989}, Particle Swarm Optimization \cite{eberhart_new_1995} or Reincarnation Algorithm (RA) \cite{sharma_new_2010} can be used to get the optimum order of bars based on their respective rank. Thereafter, a new set of images was created using this new order of bars. This process has been elaborated more in Section \ref{sec:sec6}.

\subsection{Normalized Distance matrix}\label{subsec:norm_dist_mat}
\begin{figure}[tb]
	\begin{center}
		\includegraphics[width=0.65\linewidth]{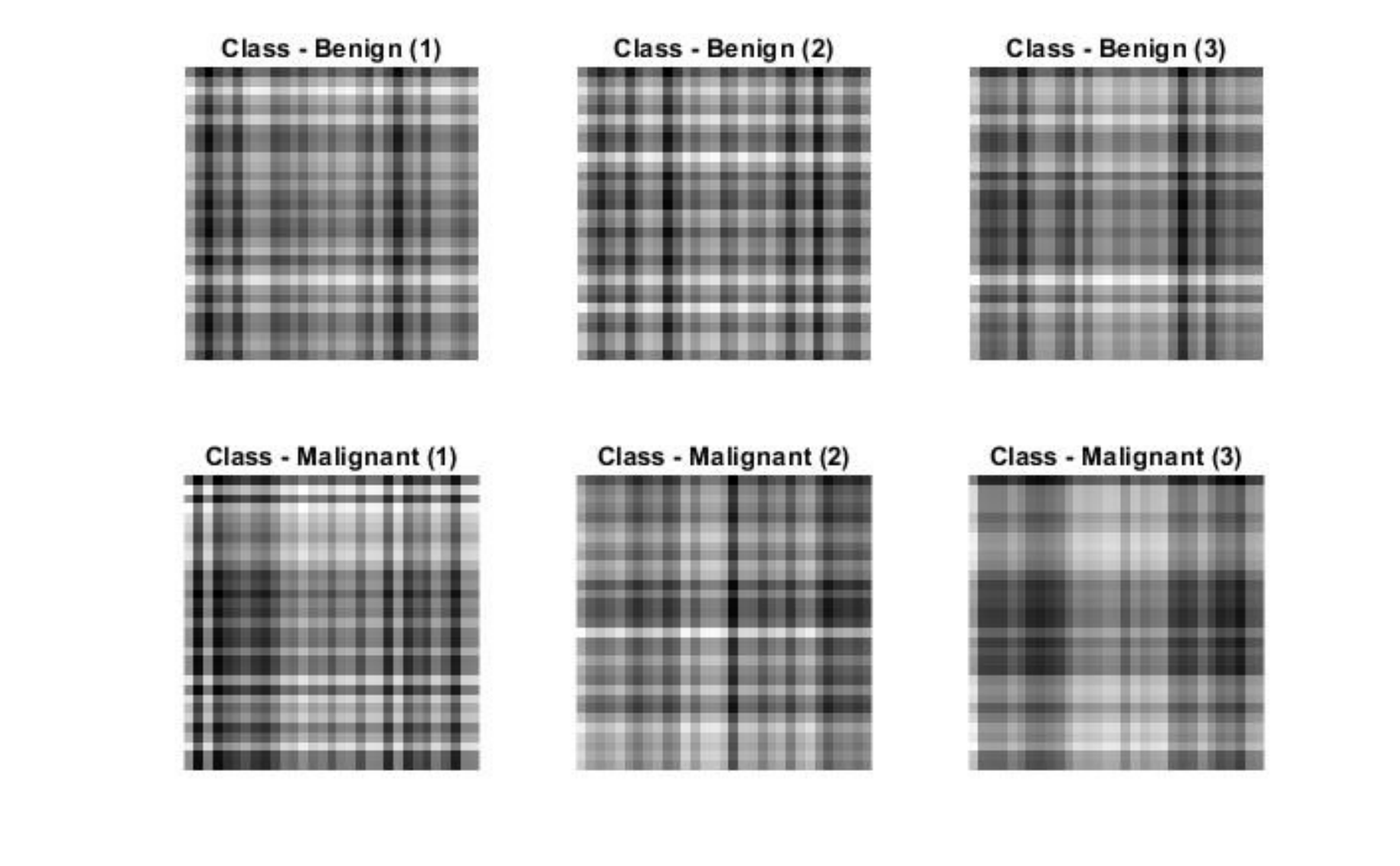}
	\end{center}
	\vspace*{-8mm}
	\caption{The normalized distance matrix for some data examples of WDBC dataset.}
	\label{fig:norm_dist_mat}
\end{figure}

\begin{figure}[tb]
	\begin{center}
		\includegraphics[width=0.75\linewidth]{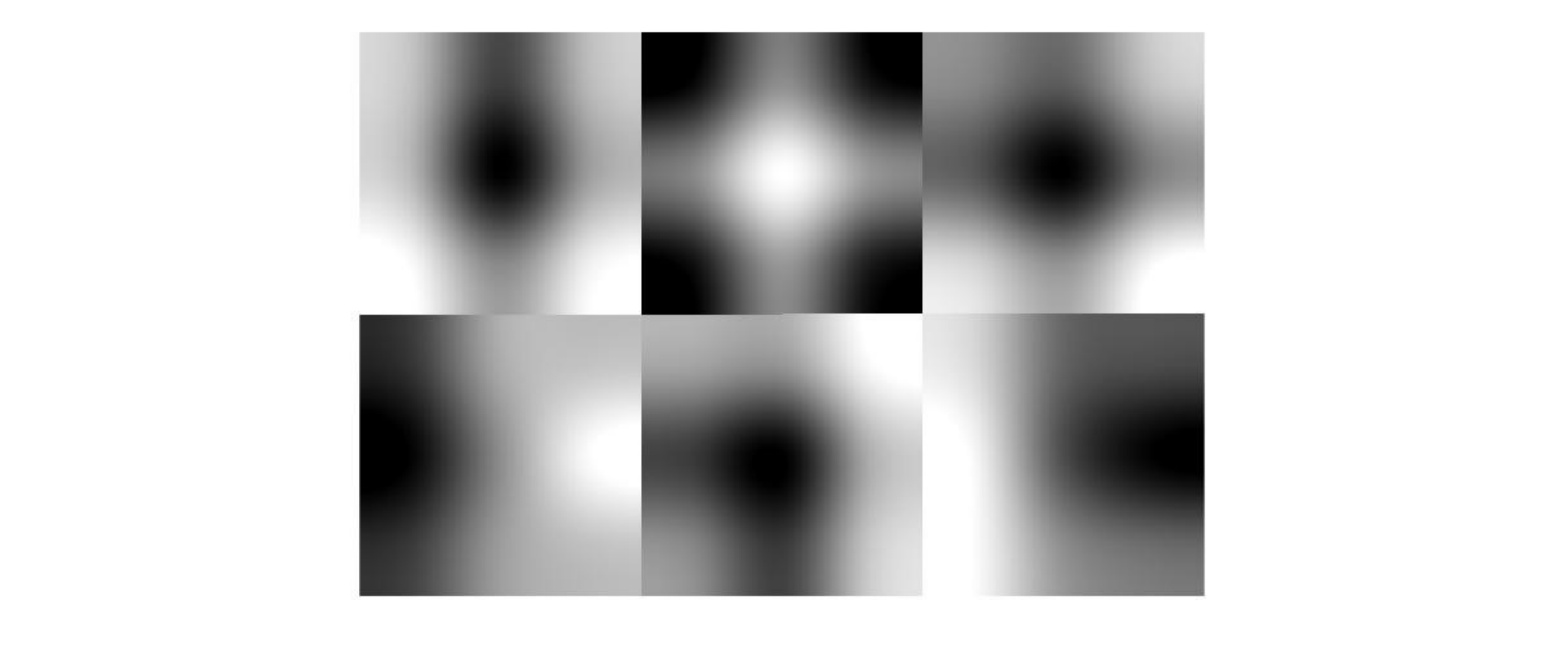}
	\end{center}
	\vspace*{-5mm}
	\caption{Features learned by the first convolutional layer for WDBC dataset with normalized distance matrix.}
	\label{fig:features_norm_dist_mat}
\end{figure}

The next method is the formation of a distance matrix which is a squared matrix of size $[d\times d]$ where $d$ represents total features of a given example. Matrix elements are the difference between two features i.e., $x_{ij}=x_i-x_j$ where $x_i$ and $x_j$ represent the measurement of a given feature with $i,j\in[1,d]$. We used Euclidean distance in our experiments. The matrix is then normalized between $[0-1]$. This produces the square image of size $[d \times d]$ which has a gain of 3 folds in length compared to bar graphs described in Section \ref{subsec:equidis}. Few data examples of WDBC dataset converted to normalized distance matrix are shown in \figurename\textbf{ }\ref{fig:norm_dist_mat} with class labels. The images can be easily scaled up to $[3d\times 3d]$. The first convolutional layer produces $6$ features similar to bar graphs is shown in \figurename\textbf{ }\ref{fig:features_norm_dist_mat}.

\subsection{Combination of options (bar graph, distance matrix, normalized numeric data)}
\begin{figure}[tb]
	\begin{center}
		\includegraphics[width=0.65\linewidth]{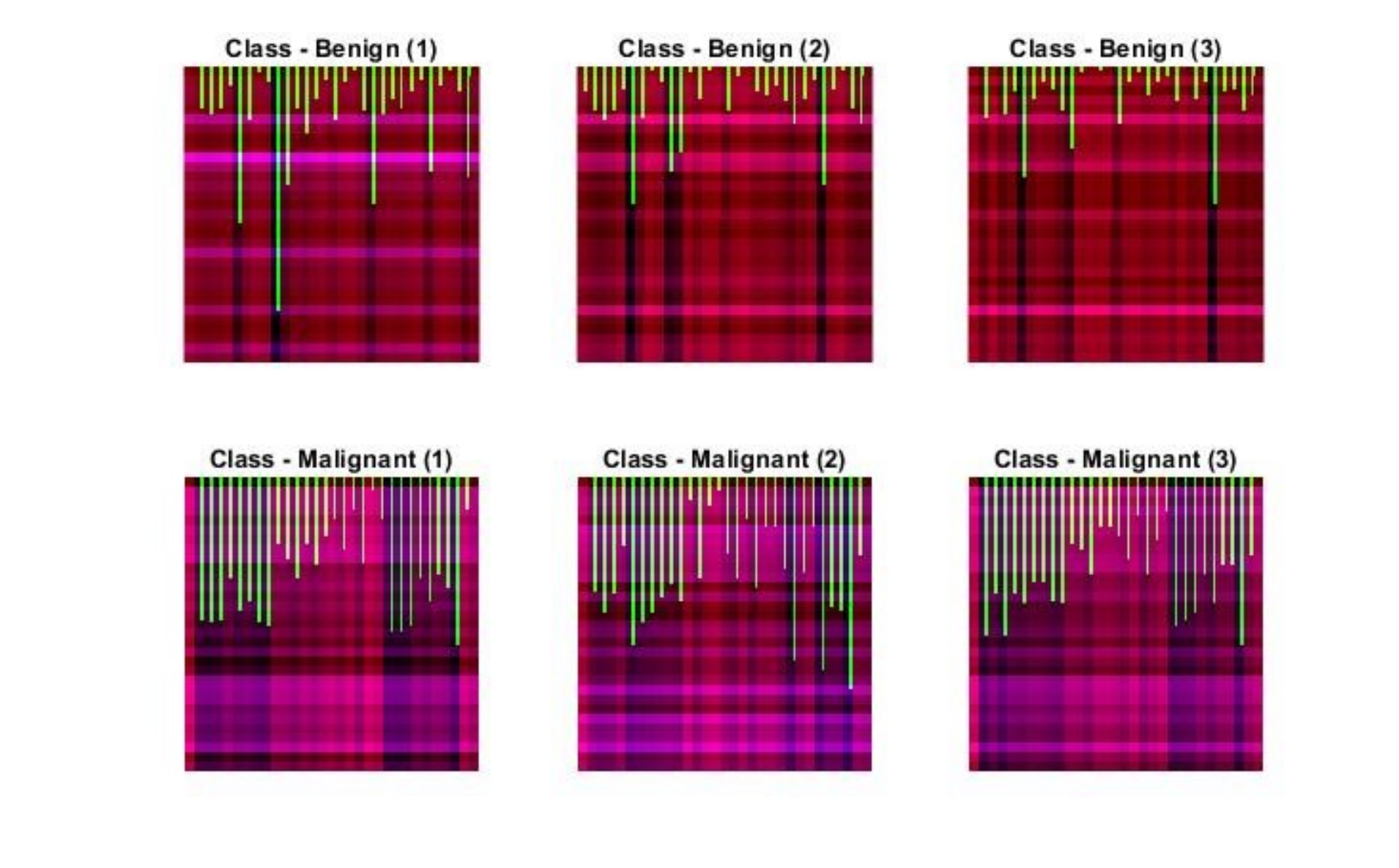}
	\end{center}
	\vspace*{-8mm}
	\caption{Combined 3 layered matrix (colored image) for some data examples of WDBC dataset.}
	\label{fig:3_layer_mat}
\end{figure}

Apparently, the above two strategies can be combined to give a third option for generating an image from numerical data. We create a colored image of 3 layers of size $[3d\times3d]$ where the first layer has a normalized distance matrix, the second layer has bar graphs, and the third layer has a copy of numerical data stored row-wise, i.e., $x_{ij}=x_i$ where $i,j\in[1,d]$ shows row and column of a matrix and $x_i$ represents the measurement of a given feature. Few data examples of WDBC dataset converted to the combination of options are shown in \figurename\textbf{ }\ref{fig:3_layer_mat} with the class labels.

\begin{figure}[tb]
	\begin{center}
		\includegraphics[width=0.65\linewidth]{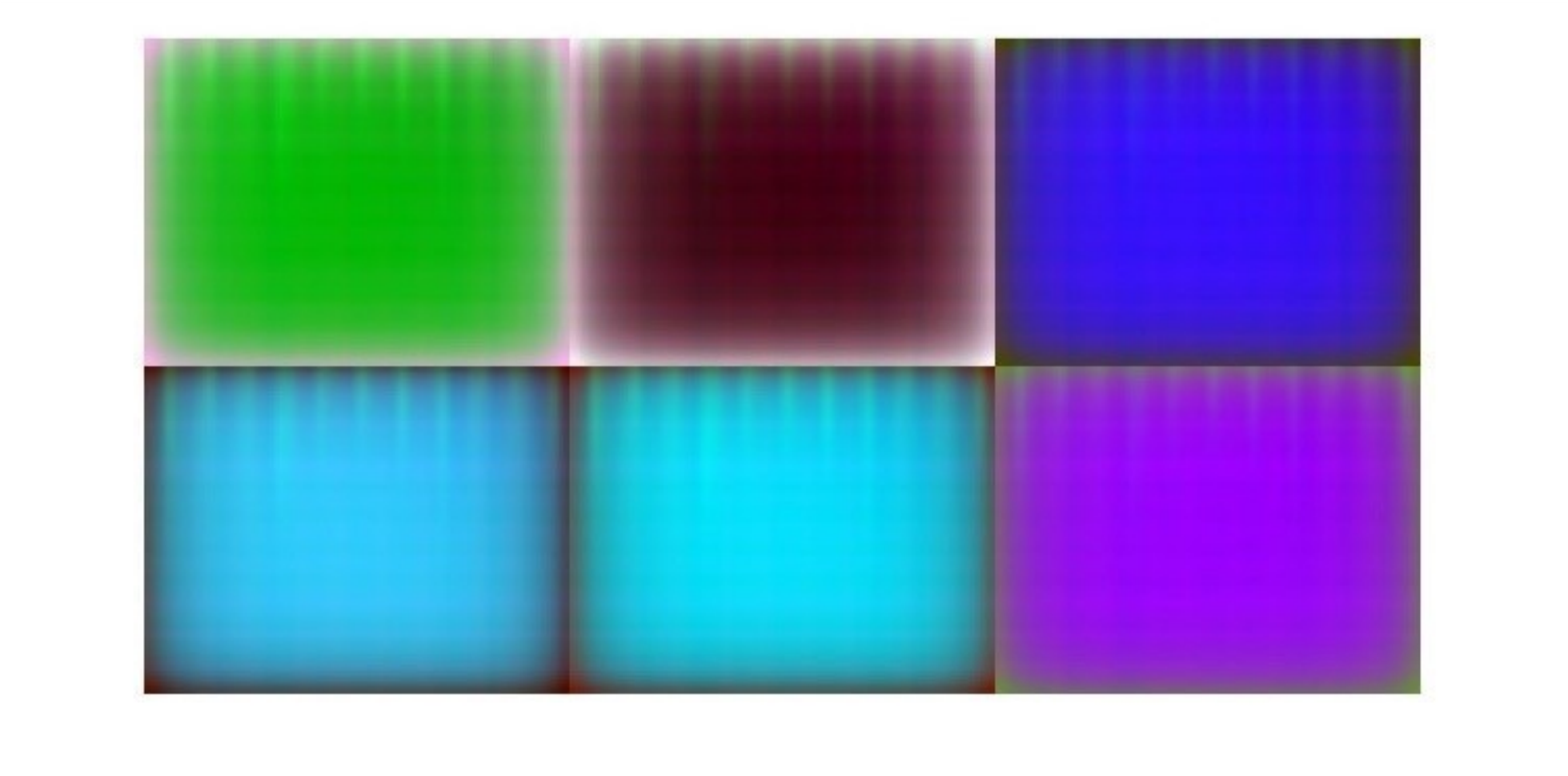}
	\end{center}
	\vspace*{-5mm}
	\caption{Features learned by the first convolutional layer for WDBC dataset with normalized distance matrix.}
	\label{fig:features_1st_Conv}
\end{figure}

\begin{figure}[tb]
	\begin{center}
		\includegraphics[width=\textwidth,height=1.5in]{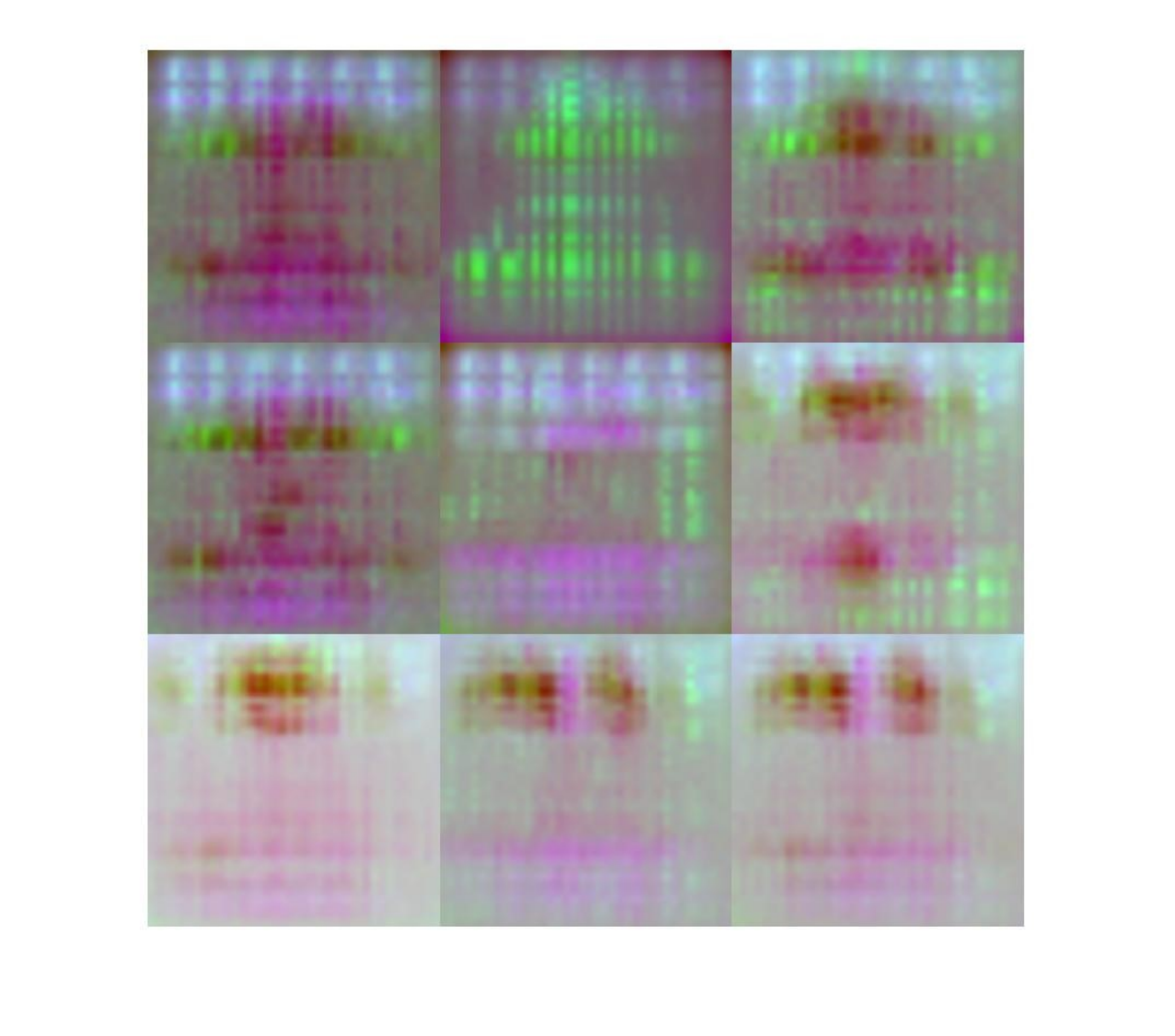}
	\end{center}
	\vspace*{-5mm}
	\caption{Features learned by the $12^{th}$ convolutional layer for WDBC dataset.}
	\label{fig:features_12th_Conv}
\end{figure}

The first convolutional layer in this case, is not able to produce any distinct feature but the scaled up image shows different colors with some bars in \figurename\textbf{ }\ref{fig:features_1st_Conv}. The $3^{rd}$ convolved block ($12^{th}$ layer) produces some blobs scattered in the images in \figurename\textbf{ }\ref{fig:features_12th_Conv}.

\section{Classification of Non-Image Data With CNN} \label{sec:sec4}
\begin{figure*}[tb]
	\begin{center}
		\includegraphics[width=1.00\linewidth]{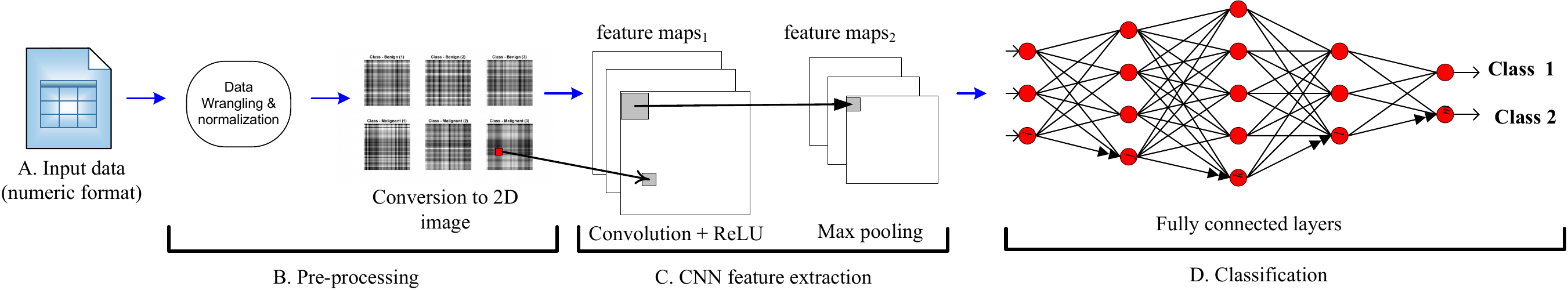}
	\end{center}
	\vspace*{-1mm}
	\caption{A complete process of non-image data classification with CNN.}
	\label{fig:flowchart}
\end{figure*}

As described in Section \ref{sec:sec2}, CNN completes the classification process in two steps. The first step is the auto-feature extraction of the images and the second step is the classification of the same images with backpropagation neural networks. In the case of a numerical dataset that is not in the form of images, first goes through the data wrangling process described in Section \ref{sec:sec3}, where either of the three options is used for non-image to image data conversion. The transformed images may not make logical sense to human eyes but CNN is capable to extract relevant features out of it. \figurename\textbf{ }\ref{fig:flowchart} illustrates the complete flowchart of the training process of CNN with non-image data sets. The process contains four important parts: Firstly, numeric input data (A) undergoes pre-processing of data wrangling (B) where it is normalized and converted to 2D image format using one of the data wrangling techniques described in Section \ref{sec:sec3} (the figure shows distance matrix method of Section \ref{subsec:norm_dist_mat}). The generated image is filtered through the CNN convolution layers for feature extraction (C). The features are trained in the fully connected layers to obtain classification outputs (D).

\section{Experiments}\label{sec:sec5}
\begin{sloppypar}
The objective of the experiment is to provide an alternative classification method with CNN for the non-image dataset of Breast Cancer and other similar datasets without any need for manual feature selection. We have used WBC and WDBC datasets from the UCI library \cite{dheeru_uci_2019} for the experiments. The properties of these datasets are given in Table \ref{tbl:datasets}. We have tested the efficacy of our method with other published state-of-the-art methods used for Breast Cancer diagnosis, namely, variations of Neural Networks (NN) \cite{bhardwaj_breast_2015}, Support Vector Machine (SVM) \cite{chen_support_2011,liu_novel_2019, zheng_breast_2014}, Decision Tree (DT) \cite{liu_decision_2009} and Na\"ive Bayes (NB) \cite{karabatak_new_2015}. These methods are generally supported by additional feature selection methods such as IG, Rough set or weight NB. 

\begin{table}[tbhp]
\caption{Experimented Dataset}
\makebox[\linewidth]{
\begin{tabular}{|c|c|c|c|c|}
\hline
\rowcolor[HTML]{EFEFEF} 
\textbf{Dataset}             & \multicolumn{1}{l|}{\cellcolor[HTML]{EFEFEF}\textbf{Attributes}} & \multicolumn{1}{l|}{\cellcolor[HTML]{EFEFEF}\textbf{Instances}} & \multicolumn{1}{l|}{\cellcolor[HTML]{EFEFEF}\textbf{Missing Values}} & \textbf{\begin{tabular}[c]{@{}c@{}}Class Ratio \\ (Benign:Malignant)\end{tabular}} \\ \hline
\cellcolor[HTML]{EFEFEF}WDBC & 32                                                               & 569                                                             & 0                                                                    & 357:212                                                                             \\ \hline
\cellcolor[HTML]{EFEFEF}WBC  & 10                                                               & 699                                                             & 16                                                                   & 458:241                                                                             \\ \hline
\end{tabular}
}
\label{tbl:datasets}
\end{table}

For CNN, we used VGG16 \cite{simonyan_very_2014} architecture with 4 convolutional blocks. Each convolutional block has 2D convolutional layer with the filter size of $[3\times3]$, $0.5\times Layer\times\left|\sqrt[]{\parallel image \parallel}\right|$ filters, ReLU layer and lastly max pooling layer with of pool size and stride of $[2\times2]$. Additionally, Bayesian optimization was used for parameter tuning. All parameter settings are shown in Table \ref{tbl:params_cnn}. For regularization and initial learning rate we used log transformation. 
\end{sloppypar}

\begin{table}
\caption{{\bf Parameter setting for CNN}}
\makebox[\linewidth]{
\begin{tabular}{|l|c|} 
\hline
\rowcolor[rgb]{0.937,0.937,0.937} \textbf{Parameter}    & \textbf{Value}   \\ 
\hline
Max iterations                                          & 1000             \\ 
\hline
Attempts                                                & 30               \\ 
\hline
Filter size                                             & 3$\times$3       \\ 
\hline
Initial learning rate $\eta$ (with log transformation)  & 0.02             \\ 
\hline
Momentum                                                & 0.88             \\ 
\hline
L2 regularization                                       & 9.4E-7           \\ 
\hline
Batch Size                                              & 8                \\
\hline
\end{tabular}
}
\label{tbl:params_cnn}
\end{table}

Initially, every dataset is divided into $80\%$ training and $20\%$ testing then $20\%$ of training data is kept aside for validation data. After 30 attempts on each dataset, we have collected best and average classification accuracies on validation and test data sets shown in Tables \ref{tbl:myBest} and \ref{tbl:myAvg} respectively. Bold figures represent the overall best result. CNN types 1, 2 and 3 represent equidistant bar graph, normalized distant matrix, and combined options respectively. $px1$ shows that the image is formed with bars of 1-pixel width only. Similarly $px2$ and $px4$ show width of 2 and 4 pixel sizes respectively for bars in an image. 

Additionally, it is highly desirable in medical diagnosis to have high sensitivity and specificity measures. Sensitivity is the ability of a test to correctly identify those with the disease, and specificity is correctly identifying those without the disease. Alternatively, the F1 score can be used as a derived metric that merges both sensitivity and precision measures. Tables \ref{tbl:measures} and \ref{tbl:avgMeasures} show the best and average of these additional metrics respectively, for WDBC and WBC datasets on classification. We have also performed experiments using CNN with 1-D convolutions on raw data without any sophisticated data transformation. However, we have obtained poor results when compared to our method with the average classification accuracy of 76.11 and 89.64 for WDBC and WBC datasets respectively.

\begin{table}[tbhp]
\caption{{\bf Best results obtained on classification accuracy}}
\makebox[\linewidth]{
\begin{tabular}{|
>{\columncolor[HTML]{EFEFEF}}c |
>{\columncolor[HTML]{EFEFEF}}c |l|l|l|l|l|l|}
\hline
\cellcolor[HTML]{EFEFEF}                                   & \cellcolor[HTML]{EFEFEF}                                                                                        & \multicolumn{6}{c|}{\cellcolor[HTML]{EFEFEF}\textbf{Image Size}}                                                                                                                                                                           \\ \cline{3-8} 
\cellcolor[HTML]{EFEFEF}                                   & \cellcolor[HTML]{EFEFEF}                                                                                        & \multicolumn{2}{c|}{\cellcolor[HTML]{EFEFEF}\textbf{px1}}                    & \multicolumn{2}{c|}{\cellcolor[HTML]{EFEFEF}\textbf{px2}}                    & \multicolumn{2}{c|}{\cellcolor[HTML]{EFEFEF}\textbf{px4}}                    \\ \cline{3-8} 
\multirow{-3}{*}{\cellcolor[HTML]{EFEFEF}\textbf{Dataset}} & \multirow{-3}{*}{\cellcolor[HTML]{EFEFEF}\textbf{\begin{tabular}[c]{@{}c@{}}Transforation\\ Type\end{tabular}}} & \cellcolor[HTML]{EFEFEF}\textbf{Val} & \cellcolor[HTML]{EFEFEF}\textbf{Test} & \cellcolor[HTML]{EFEFEF}\textbf{Val} & \cellcolor[HTML]{EFEFEF}\textbf{Test} & \cellcolor[HTML]{EFEFEF}\textbf{Val} & \cellcolor[HTML]{EFEFEF}\textbf{Test} \\ \hline
\cellcolor[HTML]{EFEFEF}                                   & 1                                                                                                               & 100.00                               & 99.12                                 & 98.90                                & 97.35                                 & 98.90                                & 98.23                                 \\ \cline{2-8} 
\cellcolor[HTML]{EFEFEF}                                   & 2                                                                                                               & 98.90                                & 94.69                                 & 97.80                                & 96.46                                 & 97.80                                & 97.35                                 \\ \cline{2-8} 
\multirow{-3}{*}{\cellcolor[HTML]{EFEFEF}WDBC}             & 3                                                                                                               & 98.90                                & \textbf{100.00}                       & 98.90                                & 99.12                                 & 100.00                               & 98.23                                 \\ \hline
\cellcolor[HTML]{EFEFEF}                                   & 1                                                                                                               & 100.00                               & 98.54                                 & 100.00                               & 98.54                                 & 100.00                               & \textbf{99.27}                                 \\ \cline{2-8} 
\cellcolor[HTML]{EFEFEF}                                   & 2                                                                                                               & 100.00                               & 97.08                                 & 99.08                                & \textbf{99.27}                                 & 98.17                                & 97.81                                 \\ \cline{2-8} 
\multirow{-3}{*}{\cellcolor[HTML]{EFEFEF}WBC}              & 3                                                                                                               & 100.00                               & \textbf{99.27}                                 & 100.00                               & 98.54                                 & 100.00                               & \textbf{99.27}                                 \\ \hline
\end{tabular}
}
\label{tbl:myBest}
\end{table}

\begin{table}[tbhp]
\caption{{\bf Average results for classification accuracy}}
\begin{tabular}{|
>{\columncolor[HTML]{EFEFEF}}c |
>{\columncolor[HTML]{EFEFEF}}c |l|l|l|l|l|l|}
\hline
\cellcolor[HTML]{EFEFEF}                                   & \cellcolor[HTML]{EFEFEF}                                                                                        & \multicolumn{6}{c|}{\cellcolor[HTML]{EFEFEF}\textbf{Image Size}}                                                                                                                                                                           \\ \cline{3-8} 
\cellcolor[HTML]{EFEFEF}                                   & \cellcolor[HTML]{EFEFEF}                                                                                        & \multicolumn{2}{c|}{\cellcolor[HTML]{EFEFEF}\textbf{px1}}                    & \multicolumn{2}{c|}{\cellcolor[HTML]{EFEFEF}\textbf{px2}}                    & \multicolumn{2}{c|}{\cellcolor[HTML]{EFEFEF}\textbf{px4}}                    \\ \cline{3-8} 
\multirow{-3}{*}{\cellcolor[HTML]{EFEFEF}\textbf{Dataset}} & \multirow{-3}{*}{\cellcolor[HTML]{EFEFEF}\textbf{\begin{tabular}[c]{@{}c@{}}Transforation\\ Type\end{tabular}}} & \cellcolor[HTML]{EFEFEF}\textbf{Val} & \cellcolor[HTML]{EFEFEF}\textbf{Test} & \cellcolor[HTML]{EFEFEF}\textbf{Val} & \cellcolor[HTML]{EFEFEF}\textbf{Test} & \cellcolor[HTML]{EFEFEF}\textbf{Val} & \cellcolor[HTML]{EFEFEF}\textbf{Test} \\ \hline
\cellcolor[HTML]{EFEFEF}                                   & 1                                                                                                               & 96.37                                & 95.19                                 & 96.81                                & 95.25                                 & 96.56                                & 95.87                                 \\ \cline{2-8} 
\cellcolor[HTML]{EFEFEF}                                   & 2                                                                                                               & 93.99                                & 91.21                                 & 92.60                                & 91.47                                 & 93.19                                & 91.95                                 \\ \cline{2-8} 
\multirow{-3}{*}{\cellcolor[HTML]{EFEFEF}WDBC}             & 3                                                                                                               & 96.70                                & \textbf{96.02}                        & 96.15                                & 95.01                                 & 96.19                                & 95.07                                 \\ \hline
\cellcolor[HTML]{EFEFEF}                                   & 1                                                                                                               & 97.22                                & 95.99                                 & 97.71                                & 96.23                                 & 97.71                                & 95.55                                 \\ \cline{2-8} 
\cellcolor[HTML]{EFEFEF}                                   & 2                                                                                                               & 94.25                                & 92.77                                 & 94.31                                & 93.67                                 & 95.60                                & 94.11                                 \\ \cline{2-8} 
\multirow{-3}{*}{\cellcolor[HTML]{EFEFEF}WBC}              & 3                                                                                                               & 97.06                                & 96.40                                 & 97.49                                & \textbf{96.55}                        & 97.19                                & 96.08                                 \\ \hline
\end{tabular}
\label{tbl:myAvg}
\end{table}

\begin{table}[]
\caption{{\bf Best Score with Type3 on px1}}
\makebox[\linewidth]{
\begin{tabular}{|c|c|c|c|l|c|}
\hline
\rowcolor[HTML]{EFEFEF} 
\cellcolor[HTML]{EFEFEF}                                   & \cellcolor[HTML]{EFEFEF}                                      & \multicolumn{4}{c|}{\cellcolor[HTML]{EFEFEF}\textbf{Score}}                                                                                                                                                            \\ \cline{3-6} 
\rowcolor[HTML]{EFEFEF} 
\multirow{-2}{*}{\cellcolor[HTML]{EFEFEF}\textbf{Dataset}} & \multirow{-2}{*}{\cellcolor[HTML]{EFEFEF}\textbf{Score Type}} & \multicolumn{1}{l|}{\cellcolor[HTML]{EFEFEF}\textbf{Sensitivity}} & \multicolumn{1}{l|}{\cellcolor[HTML]{EFEFEF}\textbf{Specificity}} & \textbf{F1} & \multicolumn{1}{l|}{\cellcolor[HTML]{EFEFEF}\textbf{Time (sec)}} \\ \hline
\cellcolor[HTML]{EFEFEF}                                   & \cellcolor[HTML]{EFEFEF}Best Sensitivity                      & 1.00                                                              & 1.00                                                              & 1.00        & 13.3                                                             \\ \cline{2-6} 
\multirow{-2}{*}{\cellcolor[HTML]{EFEFEF}WDBC}             & \cellcolor[HTML]{EFEFEF}Best Specificity                      & 1.00                                                              & 1.00                                                              & 1.00        & 9.8                                                              \\ \hline
\cellcolor[HTML]{EFEFEF}                                   & \cellcolor[HTML]{EFEFEF}Best Sensitivity                      & 1.00                                                              & 0.99                                                              & 0.99        & 15.9                                                             \\ \cline{2-6} 
\multirow{-2}{*}{\cellcolor[HTML]{EFEFEF}WBC}              & \cellcolor[HTML]{EFEFEF}Best Specificity                      & 0.96                                                              & 1.00                                                              & 0.98        & 12.8                                                             \\ \hline
\end{tabular}
}
\label{tbl:measures}
\end{table}

\begin{table}[]
\caption{{\bf Average Score with Type3 on px1}}
\begin{center}
    \begin{tabular}{|l|l|l|l|}
\hline
\rowcolor[HTML]{EFEFEF} 
\textbf{Dataset}                               & \textbf{Score Type} & \textbf{Avg Score} & \textbf{Run Time}          \\ \hline
\cellcolor[HTML]{EFEFEF}                       & Specificity         & 0.96               &                            \\ \cline{2-3}
\cellcolor[HTML]{EFEFEF}                       & Sensitivity         & 0.96               &                            \\ \cline{2-3}
\multirow{-3}{*}{\cellcolor[HTML]{EFEFEF}WDBC} & F1                  & 0.94               & \multirow{-3}{*}{13.2 sec} \\ \hline
\cellcolor[HTML]{EFEFEF}                       & Specificity         & 0.97               &                            \\ \cline{2-3}
\cellcolor[HTML]{EFEFEF}                       & Sensitivity         & 0.97               &                            \\ \cline{2-3}
\multirow{-3}{*}{\cellcolor[HTML]{EFEFEF}WBC}  & F1                  & 0.96               & \multirow{-3}{*}{13.5 sec} \\ \hline
\end{tabular}
\end{center}
\label{tbl:avgMeasures}
\end{table}

The comparison of our methods with other state-of-the-art methods is shown in Table \ref{tbl:comparison}. The table shows different methods from 2009 - 2019. The results show accuracy, sensitivity and specificity of WBC and/or WDBC datasets. Authors in \cite{sun_enhancing_2017} have used mammogram images of breast cancer as CNN works on images. In some cases, authors got 100\% accuracy with 10-fold cross-validation for WBC dataset. Lower fold of cross-validation generally gives lower accuracy \cite{chen_support_2011,liu_novel_2019,bhardwaj_breast_2015}.

\begin{table}
\caption{Comparison of the proposed method with other methods}
\makebox[\linewidth]{
\begin{tabular}{|l|l|l|l|l|l|l|} 
\hline
\rowcolor[rgb]{0.937,0.937,0.937} \multicolumn{1}{|c|}{ \textbf{Authors} } & \multicolumn{1}{c|}{\textbf{Year} } & \multicolumn{1}{c|}{\textbf{Method} } & \multicolumn{1}{c|}{\textbf{Accuracy} } & \multicolumn{1}{c|}{\textbf{Sensitivity} } & \multicolumn{1}{c|}{\textbf{Specificity} } & \multicolumn{1}{c|}{\textbf{Dataset} }  \\ 
\hline
{\cellcolor[rgb]{0.937,0.937,0.937}}Akay                                   & \multicolumn{1}{c|}{2009}           & SVM with F-score feature selection    & 99.51\%                                 & 100                                        & 97.91                                      & WBC                                     \\ 
\hline
{\cellcolor[rgb]{0.937,0.937,0.937}}Chen et al.                            & 2011                                & Rough set (RS) and SVM                & \textbf{100\%}                          & \textbf{100}                               & \textbf{100}                               & WBC                                     \\ 
\hline
{\cellcolor[rgb]{0.937,0.937,0.937}}Onan                                   & 2015                                & Fuzzy-rough nearest neighbor          & 99.72\%                                 & 100                                        & 99.47                                      & WBC                                     \\ 
\hline
{\cellcolor[rgb]{0.937,0.937,0.937}}Bhardwaj et al.                        & 2015                                & Genetically Optimized NN              & \textbf{100\%}                          & 98.77                                      & \textbf{100}                               & WBC                                     \\ 
\hline
{\cellcolor[rgb]{0.937,0.937,0.937}}Karabatak                              & 2015                                & Naïve Bayesian (NB)                   & 98.54\%                                 & 99.11                                      & 98.25                                      & WBC                                     \\ 
\hline
{\cellcolor[rgb]{0.937,0.937,0.937}}Wang et al.                            & 2018                                & SVM based ensemble learning           & 97.10\%                                 & 97.11                                      & 97.23                                      & WBC                                     \\ 
\hline
{\cellcolor[rgb]{0.937,0.937,0.937}}Na Liu et al.                          & 2019                                & IGSAGAW with CSSVM                    & 95.80\%                                 & -                                          & -                                          & WBC                                     \\ 
\hline
{\cellcolor[rgb]{0.937,0.937,0.937}}\textit{of this paper}                 & 2020                                & CNN with Type-3 Transformation        & 99.27                                   & \textbf{100}                               & 98.88                                      & WBC                                     \\ 
\hline
{\cellcolor[rgb]{0.937,0.937,0.937}}Ahn et al.                             & \multicolumn{1}{c|}{2009}           & Novel CBR                             & 99.12\%                                 & -                                          & -                                          & WDBC                                    \\ 
\hline
{\cellcolor[rgb]{0.937,0.937,0.937}}Sun et al.                             & 2017                                & \textit{CNN on mammogram images }     & \textit{82.43\% }                       & \textit{81.00 }                            & \textit{72.26 }                            & \textit{Mammogram }                     \\ 
\hline
{\cellcolor[rgb]{0.937,0.937,0.937}}Wang et al.                            & 2018                                & SVM based ensemble learning           & 97.68\%                                 & 94.75                                      & 99.49                                      & WDBC                                    \\ 
\hline
{\cellcolor[rgb]{0.937,0.937,0.937}}Na Liu et al.                          & 2019                                & IGSAGAW with CSSVM                    & 95.70\%                                 & -                                          & -                                          & WDBC                                    \\ 
\hline
{\cellcolor[rgb]{0.937,0.937,0.937}}\textit{of this paper}                 & 2020                                & CNN with Type-3 Transformation        & \textbf{100\%}                          & \textbf{100}                               & 100                                      & WDBC                                    \\
\hline
\end{tabular}
}
\label{tbl:comparison}
\end{table}

\section{Discussion}\label{sec:sec6}
The experimental results of data transformation from non-image breast cancer datasets to image have been promising for the utilization of CNN for classification accuracy. Although the proposed methods are in the early stages, the obtained results are very significant in the development of new strategies with data wrangling for deep learning. This also provides an opportunity to derive even better alternatives for CNN in the future. It was observed that our proposed combined approach, i.e. Type-3 transformation and bar width of 1 pixel i.e. $px1$, has been the most significant method as it carries the most information about the data in three dimensions of an image. It has outperformed other methods for the WDBC dataset by clocking 100\% accuracy (with 1.0 sensitivity, specificity and F1 score). It has also shown very competitive results for the WBC dataset with 99.27\% accuracy and 1.0 sensitivity 0.99 specificity and 0.99 F1 score. 

\begin{figure}[tb]
	\begin{center}
		\includegraphics[width=0.75\linewidth]{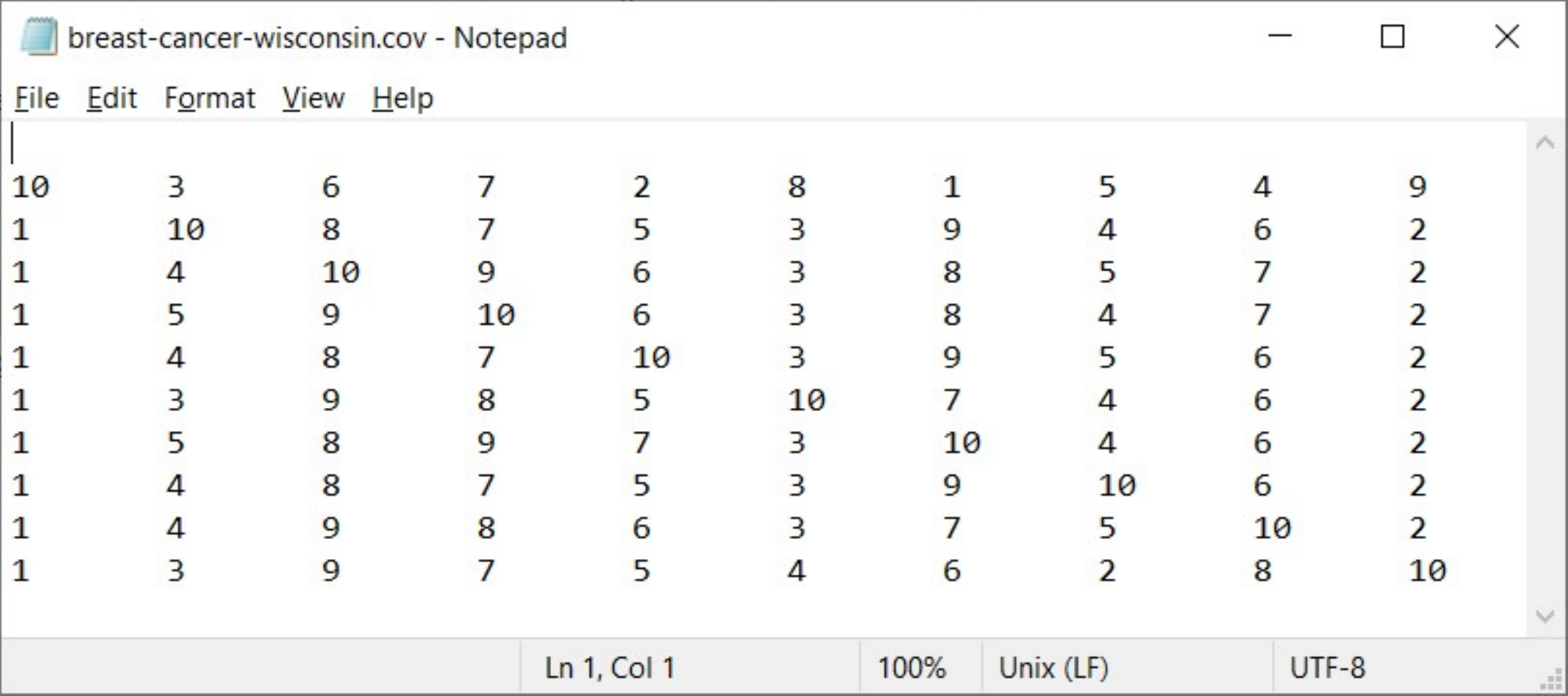}
	\end{center}
	\vspace*{-1mm}
	\caption{Ranking of co-variance for WBC dataset in Adjacency Matrix}
	\label{fig:CovarianceFile}
\end{figure}

\begin{figure}[H]
	\begin{center}
		\includegraphics[width=0.65\linewidth]{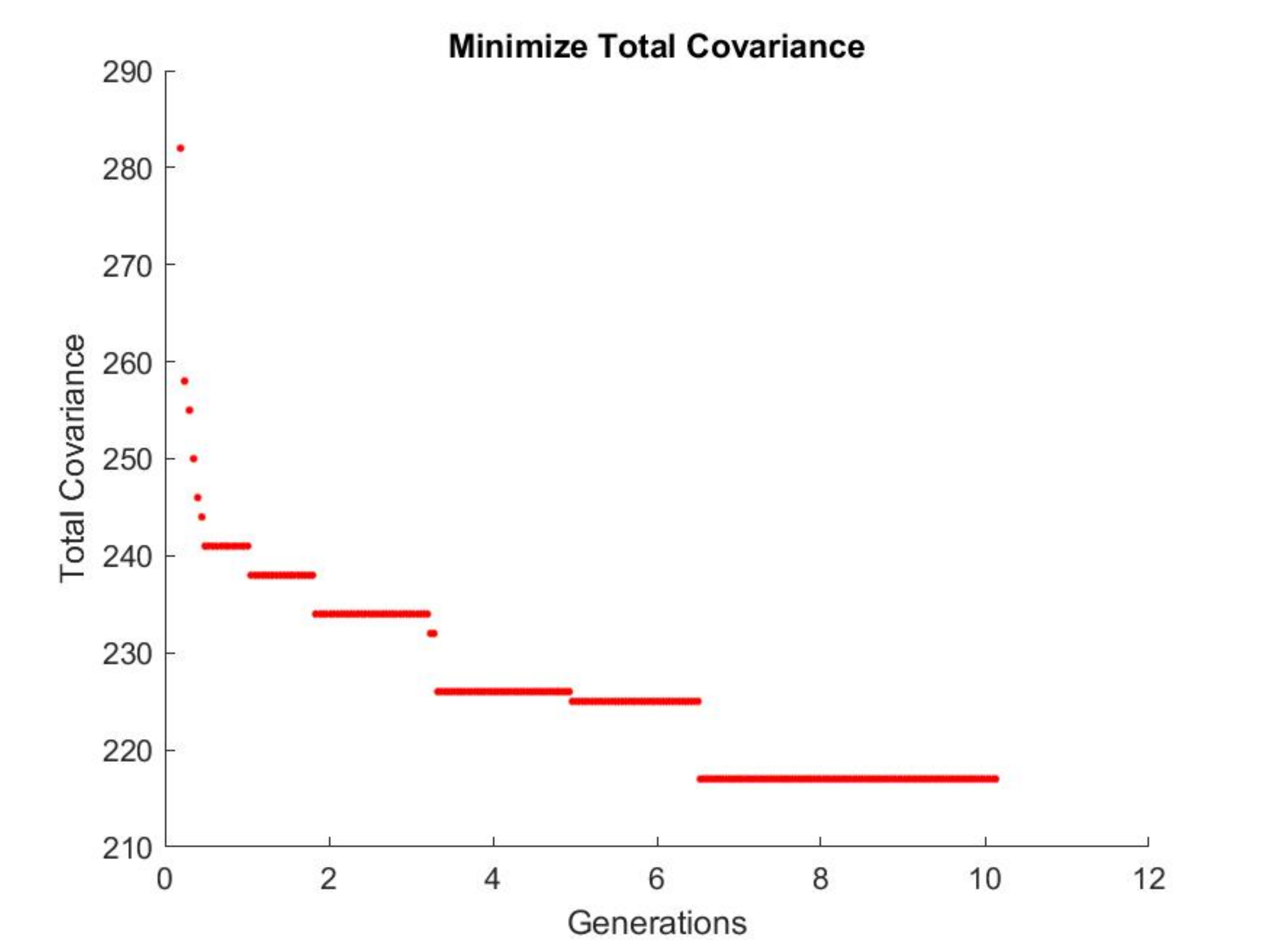}
	\end{center}
	\vspace*{-2mm}
	\caption{Minimization of total covariance for a given combination of fields for WDBC dataset}
	\label{fig:GAforBreast}
\end{figure}

As discussed in Section \ref{sec:sec3}, different order of bar graphs for Type-1 and Type-3 transformations produces different images. A bar represents its corresponding field value of a given sample. We have tried to bring the related bars closer to each other by using a covariance matrix that determines the \enquote{closeness} of two fields. For example \figurename\textbf{ }\ref{fig:CovarianceFile} shows the Adjacency Matrix of co-variance of each field for WBC dataset. The data is arranged row-wise such that each value represents the rank of $i^{th}$ row with $j^{th}$ column of a given field. To get the ``best" arrangement of fields, we minimize the total co-variance rank by using a meta-heuristic algorithm GA to solve this shortest path problem. The process of minimization for WDBC is shown in \figurename\textbf{ }\ref{fig:GAforBreast} where the minimum rank is obtained by the end of $10^{th}$ generation. The dataset fields were reorganized where the related fields were put close to each other according to the order of their similarity. The final order of fields for WBC and WDBC produced through minimum ranks are shown in Table \ref{tbl:GAorder}. The images of these datasets were generated accordingly for the experiment. 

\begin{table}[]
\caption{Order of fields based on minimization of total co-variance of adjacency matrix} 
\begin{center}
\begin{tabular}{|l|l|}
\hline
\rowcolor[HTML]{EFEFEF} 
\textbf{Dataset} & \textbf{Order of Fields}                                                                                                                                           \\ \hline
WBC              & {[}5, 4, 6, 2, 3, 7, 9, 1, 10, 8{]}                                                                                                                                \\ \hline
WDBC             & \begin{tabular}[c]{@{}l@{}}{[}5, 27, 14, 16, 4, 11, 2, 10, 3, 6, 1, 7, 13, 29, 20, 24, 8, \\ 21, 22, 17, 25, 26, 12, 30, 9, 18, 23, 19, 28, 15, 31{]}\end{tabular} \\ \hline
\end{tabular}
\end{center}
\label{tbl:GAorder}
\end{table}

The only shortcoming of the CNN algorithm is its high processing cost than other methods, especially with bigger sized images. Generally, it takes 9-15 seconds for a MATLAB 2018 program to complete the training process on DELL XPS i7-9700 @ 3GHz machine with 8 CPUs and NVIDIA GEFORCE RTX 2060 GPU. Despite this, the experimental results demonstrate the size of data has no direct impact on the performance of CNN. Additionally, with the advent of quantum computing \cite{arute_quantum_2019} and parallel GPUs with enough memory can produce results in a reasonable time frame. The data wrangling process of converting non-image data to the image is not too expensive either. The every-case time complexity of the bar graph approach has the order of $O(Nd)$ and the normalized distance matrix has the order of $O(Nd^2)$. The details of the algorithms are given in the Appendix.

\section{Conclusion}\label{sec:sec7}
The objective of this paper was to process non-image data (in a non-time series form) of Breast Cancer datasets WDBC and WBC into CNN due to its state-of-the-art performance and elimination of manual feature extraction for image recognition applications. The utilization of CNN has been confined largely to image data only except for some domain-specific data conversion techniques such as NLP and voice recognition. We have proposed some novel approaches to convert numerical non-time series data to image data. This process of conversion is very straightforward with the efficiency of the order of not more than $O(Nd^2)$. The experimental results on classification accuracy show the competitiveness of these methods. There is also a high potential for improving these approaches further to have more outstanding results. For example, bar graphs with different shapes, sizes, color and even arrangements can be tried. Similarly, distance matrix can be enhanced to have more information such as the mean/variance of the neighboring elements. It still needs to be seen how other applications with various types and orientations of numerical data would respond to CNN after non-image data conversion to image data. Intuitively, the more the information on data would produce the better the results as observed with the combined approach. Finally, the classification accuracy of numerical data without any sophisticated data transformation on 1-D CNN did not produce acceptable results.    

\appendix

\section{Algorithm for Equidistant Bar Graph}\label{app:A}
The pseudocode of the algorithm for equidistant bar graph is given in Algorithm \ref{alg:algo1}. Depending on the required size of the image the parameter $\psi$ and $\gamma$ can be set to define the width and the constant gap size between two consecutive bars respectively. All the distances are in pixels. The length and width are calculated as $R$ and C. The maximum height of the bars is $H$ which leaves some padding distance. $I$ is a $0$ matrix of size $R\times C$. $X_i$ is the $i^{th}$ data example from the dataset $X$. $B_i$ is the height of bars of a data example $X_i$.

\begin{algorithm}
	$\psi \leftarrow 1;$\\
	$\gamma \leftarrow 2;$\\
	$R \leftarrow \psi * d + \gamma * (d+1);$\\
	$C \leftarrow \psi * d + \gamma * (d+1);$\\
	$R \leftarrow R - 2\psi;$\\
	$I \leftarrow O_{R\times C};$ // \textbf{0} matrix of size $R\times C$\\
	
	\For{$i = 1:N$}
	{
		$M_i \leftarrow I$\\
		$B_i \leftarrow \lfloor H *  X_i \rfloor$ //bars \\
		$J \leftarrow \gamma + 1$\\
		$k \leftarrow 1$\\
		\While{$j \leq C - \gamma$}
		{
			$G = 0;$\\
			$M_i[\psi$...$B_i(k),j$...$(j + \psi -1)] = 1$\\
			$k \leftarrow k + 1$\\
			$j \leftarrow j + \gamma + \psi$ \\
			\If{$k > d$}
			{
				$break$
			}
		}
		Save $(M_i)$
	}
	\vspace{3mm}
	\caption{Equidistant Bar Graph}
	\label{alg:algo1}	
\end{algorithm}

\section{Algorithm for Normalized Distance matrix}
The pseudocode of the algorithm for the normalized distance matrix is given in Algorithm \ref{alg:algo2}. Here we use the same parameters as used in the above section of the algorithm for Equidistant Bar Graph. Additionally, the graph can be expanded by $[e1\times e2]$ with matrix $E$ of size $[e1\times e2]$ where each element is $1$. Normalization of values between 0-1 is given by $normalize01()$ function.

\begin{algorithm}
	$E \leftarrow 
	\begin{bmatrix}
		1 & \dots & 1 \\
		\vdots & \ddots & \vdots \\
		1 & \dots & 1
	\end{bmatrix}$ //Expand by $[e1\times e2]$\\ 
	\vspace{1mm}
	$I \leftarrow O_{R\times C};$ // matrix of size $R\times C$\\
	
	\For{$i = 1:N$}
	{
		$M_i \leftarrow I$\\
		\For{$r = 1:d$}
		{
			\For{$c = 1:d$}
			{
				$M_i(r,c) = X_i(r) - X_i(c);$\\
				$M_i(r,c) = M_i(r,c) * E;$
			}
		}
		$M_i = normalize01(M_i);$\\
		Save $(M_i)$
	}
	\vspace{3mm}
	\caption{Normalized Distance Matrix}
	\label{alg:algo2}	
\end{algorithm}





\bibliographystyle{model1-num-names}
\bibliography{main.bib}

\begin{thebibliography}{62}
\expandafter\ifx\csname natexlab\endcsname\relax\def\natexlab#1{#1}\fi
\providecommand{\bibinfo}[2]{#2}
\ifx\xfnm\relax \def\xfnm[#1]{\unskip,\space#1}\fi
\bibitem[{Sourla et~al.(2012)Sourla, Sioutas, Syrimpeis, Tsakalidis, and
  Tzimas}]{sourla2012cardiosmart365}
\bibinfo{author}{E.~Sourla}, \bibinfo{author}{S.~Sioutas},
  \bibinfo{author}{V.~Syrimpeis}, \bibinfo{author}{A.~Tsakalidis},
  \bibinfo{author}{G.~Tzimas},
\newblock \bibinfo{title}{Cardiosmart365: artificial intelligence in the
  service of cardiologic patients},
\newblock \bibinfo{journal}{Advances in Artificial Intelligence}
  \bibinfo{volume}{2012} (\bibinfo{year}{2012}) \bibinfo{pages}{2}.
\bibitem[{Gao et~al.(2018)Gao, Wu, Li, Zheng, Ruan, Shang, and
  Patel}]{gao2018sd}
\bibinfo{author}{F.~Gao}, \bibinfo{author}{T.~Wu}, \bibinfo{author}{J.~Li},
  \bibinfo{author}{B.~Zheng}, \bibinfo{author}{L.~Ruan},
  \bibinfo{author}{D.~Shang}, \bibinfo{author}{B.~Patel},
\newblock \bibinfo{title}{Sd-cnn: A shallow-deep cnn for improved breast cancer
  diagnosis},
\newblock \bibinfo{journal}{Computerized Medical Imaging and Graphics}
  \bibinfo{volume}{70} (\bibinfo{year}{2018}) \bibinfo{pages}{53--62}.
\bibitem[{Tsai et~al.(2020)Tsai, Knaack, Martone, Krueger, Baldinger, Lillemoe,
  Susnik, Grimm, Olet, Rueth et~al.}]{tsai2020breast}
\bibinfo{author}{M.~L. Tsai}, \bibinfo{author}{M.~Knaack},
  \bibinfo{author}{P.~Martone}, \bibinfo{author}{J.~Krueger},
  \bibinfo{author}{S.~R. Baldinger}, \bibinfo{author}{T.~J. Lillemoe},
  \bibinfo{author}{B.~Susnik}, \bibinfo{author}{E.~Grimm},
  \bibinfo{author}{S.~Olet}, \bibinfo{author}{N.~Rueth}, et~al.,
\newblock \bibinfo{title}{Breast cancer diagnosed in young women≤ age 35:
  Effects of germline pathogenic variants, cancer subtypes, tumor-related
  characteristics, and pregnancy-associated diagnosis on outcomes},
\newblock \bibinfo{journal}{Clinical Breast Cancer}  (\bibinfo{year}{2020}).
\bibitem[{noa(????{\natexlab{a}})}]{noauthor_breast_nodate}
\bibinfo{title}{Breast cancer - {Latest} research and news {\textbar}
  {Nature}}, ????{\natexlab{a}}.
\bibitem[{noa(????{\natexlab{b}})}]{noauthor_breast_nodate-1}
\bibinfo{title}{Breast cancer {\textbar} definition of breast cancer by
  {Medical} dictionary}, ????{\natexlab{b}}.
\bibitem[{Kaur et~al.(2019)Kaur, Porras, Ring, Carpten, and
  Lang}]{kaur2019comparison}
\bibinfo{author}{P.~Kaur}, \bibinfo{author}{T.~B. Porras},
  \bibinfo{author}{A.~Ring}, \bibinfo{author}{J.~D. Carpten},
  \bibinfo{author}{J.~E. Lang},
\newblock \bibinfo{title}{Comparison of tcga and genie genomic datasets for the
  detection of clinically actionable alterations in breast cancer},
\newblock \bibinfo{journal}{Scientific reports} \bibinfo{volume}{9}
  (\bibinfo{year}{2019}) \bibinfo{pages}{1--15}.
\bibitem[{Larsen et~al.(2014)Larsen, Thomassen, Tan, S{\o}rensen, and
  Kruse}]{larsen2014microarray}
\bibinfo{author}{M.~J. Larsen}, \bibinfo{author}{M.~Thomassen},
  \bibinfo{author}{Q.~Tan}, \bibinfo{author}{K.~P. S{\o}rensen},
  \bibinfo{author}{T.~A. Kruse},
\newblock \bibinfo{title}{Microarray-based rna profiling of breast cancer:
  batch effect removal improves cross-platform consistency},
\newblock \bibinfo{journal}{BioMed research international}
  \bibinfo{volume}{2014} (\bibinfo{year}{2014}).
\bibitem[{Dembrower et~al.(2019)Dembrower, Lindholm, and
  Strand}]{dembrower2019multi}
\bibinfo{author}{K.~Dembrower}, \bibinfo{author}{P.~Lindholm},
  \bibinfo{author}{F.~Strand},
\newblock \bibinfo{title}{A multi-million mammography image dataset and
  population-based screening cohort for the training and evaluation of deep
  neural networks—the cohort of screen-aged women (csaw)},
\newblock \bibinfo{journal}{Journal of digital imaging}  (\bibinfo{year}{2019})
  \bibinfo{pages}{1--6}.
\bibitem[{Bowyer et~al.(1996)Bowyer, Kopans, Kegelmeyer, Moore, Sallam, Chang,
  and Woods}]{bowyer1996digital}
\bibinfo{author}{K.~Bowyer}, \bibinfo{author}{D.~Kopans},
  \bibinfo{author}{W.~Kegelmeyer}, \bibinfo{author}{R.~Moore},
  \bibinfo{author}{M.~Sallam}, \bibinfo{author}{K.~Chang},
  \bibinfo{author}{K.~Woods},
\newblock \bibinfo{title}{The digital database for screening mammography},
\newblock in: \bibinfo{booktitle}{Third international workshop on digital
  mammography}, volume~\bibinfo{volume}{58}, p.~\bibinfo{pages}{27}.
\bibitem[{Dheeru and Karra~Taniskidou(2019)}]{dheeru_uci_2019}
\bibinfo{author}{D.~Dheeru}, \bibinfo{author}{E.~Karra~Taniskidou},
  \bibinfo{title}{{UCI} {Machine} {Learning} {Repository}},
  \bibinfo{publisher}{University of California, Irvine, School of Information
  and Computer Sciences}, \bibinfo{year}{2019}.
\bibitem[{Sun et~al.(2017)Sun, Tseng, Zhang, and Qian}]{sun_enhancing_2017}
\bibinfo{author}{W.~Sun}, \bibinfo{author}{T.-L.~B. Tseng},
  \bibinfo{author}{J.~Zhang}, \bibinfo{author}{W.~Qian},
\newblock \bibinfo{title}{Enhancing deep convolutional neural network scheme
  for breast cancer diagnosis with unlabeled data},
\newblock \bibinfo{journal}{Computerized Medical Imaging and Graphics}
  \bibinfo{volume}{57} (\bibinfo{year}{2017}) \bibinfo{pages}{4--9}.
\bibitem[{Firmino et~al.(2016)Firmino, Angelo, Morais, Dantas, and
  Valentim}]{firmino_computer-aided_2016}
\bibinfo{author}{M.~Firmino}, \bibinfo{author}{G.~Angelo},
  \bibinfo{author}{H.~Morais}, \bibinfo{author}{M.~R. Dantas},
  \bibinfo{author}{R.~Valentim},
\newblock \bibinfo{title}{Computer-aided detection ({CADe}) and diagnosis
  ({CADx}) system for lung cancer with likelihood of malignancy},
\newblock \bibinfo{journal}{BioMedical Engineering OnLine} \bibinfo{volume}{15}
  (\bibinfo{year}{2016}) \bibinfo{pages}{2}.
\bibitem[{Guyon and Elisseeff(2003)}]{guyon2003introduction}
\bibinfo{author}{I.~Guyon}, \bibinfo{author}{A.~Elisseeff},
\newblock \bibinfo{title}{An introduction to variable and feature selection},
\newblock \bibinfo{journal}{Journal of machine learning research}
  \bibinfo{volume}{3} (\bibinfo{year}{2003}) \bibinfo{pages}{1157--1182}.
\bibitem[{Kumar and Minz(2014)}]{kumar2014feature}
\bibinfo{author}{V.~Kumar}, \bibinfo{author}{S.~Minz},
\newblock \bibinfo{title}{Feature selection: a literature review},
\newblock \bibinfo{journal}{SmartCR} \bibinfo{volume}{4} (\bibinfo{year}{2014})
  \bibinfo{pages}{211--229}.
\bibitem[{Fodor(2002)}]{fodor2002survey}
\bibinfo{author}{I.~K. Fodor}, \bibinfo{title}{A survey of dimension reduction
  techniques}, \bibinfo{type}{Technical Report}, Lawrence Livermore National
  Lab., CA (US), \bibinfo{year}{2002}.
\bibitem[{Liu et~al.(2019)Liu, Qi, Xu, Gao, and Liu}]{liu_novel_2019}
\bibinfo{author}{N.~Liu}, \bibinfo{author}{E.-S. Qi}, \bibinfo{author}{M.~Xu},
  \bibinfo{author}{B.~Gao}, \bibinfo{author}{G.-Q. Liu},
\newblock \bibinfo{title}{A novel intelligent classification model for breast
  cancer diagnosis},
\newblock \bibinfo{journal}{Information Processing \& Management}
  \bibinfo{volume}{56} (\bibinfo{year}{2019}) \bibinfo{pages}{609--623}.
\bibitem[{Babatunde et~al.(2014)Babatunde, Armstrong, Leng, and
  Diepeveen}]{babatunde2014genetic}
\bibinfo{author}{O.~H. Babatunde}, \bibinfo{author}{L.~Armstrong},
  \bibinfo{author}{J.~Leng}, \bibinfo{author}{D.~Diepeveen},
\newblock \bibinfo{title}{A genetic algorithm-based feature selection}
  (\bibinfo{year}{2014}).
\bibitem[{Darst et~al.(2018)Darst, Malecki, and Engelman}]{darst2018using}
\bibinfo{author}{B.~F. Darst}, \bibinfo{author}{K.~C. Malecki},
  \bibinfo{author}{C.~D. Engelman},
\newblock \bibinfo{title}{Using recursive feature elimination in random forest
  to account for correlated variables in high dimensional data},
\newblock \bibinfo{journal}{BMC genetics} \bibinfo{volume}{19}
  (\bibinfo{year}{2018}) \bibinfo{pages}{65}.
\bibitem[{Sharma and Kaur(2020)}]{sharma2020comprehensive}
\bibinfo{author}{M.~Sharma}, \bibinfo{author}{P.~Kaur},
\newblock \bibinfo{title}{A comprehensive analysis of nature-inspired
  meta-heuristic techniques for feature selection problem},
\newblock \bibinfo{journal}{Archives of Computational Methods in Engineering}
  (\bibinfo{year}{2020}) \bibinfo{pages}{1--25}.
\bibitem[{Pawlak(2012)}]{pawlak_rough_2012}
\bibinfo{author}{Z.~Pawlak}, \bibinfo{title}{Rough {Sets}: {Theoretical}
  {Aspects} of {Reasoning} about {Data}}, \bibinfo{publisher}{Springer Science
  \& Business Media}, \bibinfo{year}{2012}. \bibinfo{note}{Google-Books-ID:
  yeOoCAAAQBAJ}.
\bibitem[{Guyon and Elisseeff(2003)}]{guyon_introduction_2003}
\bibinfo{author}{I.~Guyon}, \bibinfo{author}{A.~Elisseeff},
\newblock \bibinfo{title}{An {Introduction} to {Variable} and {Feature}
  {Selection}},
\newblock \bibinfo{journal}{Journal of Machine Learning Research}
  \bibinfo{volume}{3} (\bibinfo{year}{2003}) \bibinfo{pages}{1157--1182}.
\bibitem[{Singh and SivaBalakrishnan(2015)}]{singh_feature_2015}
\bibinfo{author}{R.~K. Singh}, \bibinfo{author}{M.~SivaBalakrishnan},
\newblock \bibinfo{title}{Feature {Selection} of {Gene} {Expression} {Data} for
  {Cancer} {Classification}: {A} review},
\newblock in: \bibinfo{booktitle}{2nd {International} {Symposium} on {Big}
  {Data} and {Cloud} {Computing}}, pp. \bibinfo{pages}{52--57}.
\bibitem[{Mohamad et~al.(2004)Mohamad, Deris, Yatim, and
  Othman}]{mohamad_feature_2004}
\bibinfo{author}{M.~S. Mohamad}, \bibinfo{author}{S.~Deris},
  \bibinfo{author}{S.~M. Yatim}, \bibinfo{author}{M.~R. Othman},
\newblock \bibinfo{title}{Feature {Selection} method using genetic algorithm
  for the classification of small and high dimension data},
\newblock in: \bibinfo{booktitle}{First {International} {Symposium} on
  {Information} and {Communication} {Technologies}}.
\bibitem[{Kumar and Sharma(2019)}]{kumar_deep_2019}
\bibinfo{author}{D.~Kumar}, \bibinfo{author}{D.~Sharma},
\newblock \bibinfo{title}{Deep {Learning} in {Gene} {Expression} {Modeling}},
\newblock in: \bibinfo{booktitle}{Handbook of {Deep} {Learning}
  {Applications}}, \bibinfo{publisher}{Springer}, \bibinfo{year}{2019}, pp.
  \bibinfo{pages}{363--383}.
\bibitem[{Cui et~al.(2016)Cui, Chen, and Chen}]{cui_multi-scale_2016}
\bibinfo{author}{Z.~Cui}, \bibinfo{author}{W.~Chen}, \bibinfo{author}{Y.~Chen},
\newblock \bibinfo{title}{Multi-{Scale} {Convolutional} {Neural} {Networks} for
  {Time} {Series} {Classification}},
\newblock \bibinfo{journal}{arXiv:1603.06995 [cs]}  (\bibinfo{year}{2016}).
  \bibinfo{note}{ArXiv: 1603.06995}.
\bibitem[{Krizhevsky et~al.(2012)Krizhevsky, Sutskever, and
  Hinton}]{krizhevsky_imagenet_2012}
\bibinfo{author}{A.~Krizhevsky}, \bibinfo{author}{I.~Sutskever},
  \bibinfo{author}{G.~E. Hinton},
\newblock \bibinfo{title}{{ImageNet} {Classification} with {Deep}
  {Convolutional} {Neural} {Networks}},
\newblock in: \bibinfo{editor}{F.~Pereira}, \bibinfo{editor}{C.~J.~C. Burges},
  \bibinfo{editor}{L.~Bottou}, \bibinfo{editor}{K.~Q. Weinberger} (Eds.),
  \bibinfo{booktitle}{Advances in {Neural} {Information} {Processing} {Systems}
  25}, \bibinfo{publisher}{Curran Associates, Inc.}, \bibinfo{year}{2012}, pp.
  \bibinfo{pages}{1097--1105}.
\bibitem[{Simonyan and Zisserman(2014)}]{simonyan_very_2014}
\bibinfo{author}{K.~Simonyan}, \bibinfo{author}{A.~Zisserman},
\newblock \bibinfo{title}{Very {Deep} {Convolutional} {Networks} for
  {Large}-{Scale} {Image} {Recognition}},
\newblock \bibinfo{journal}{arXiv:1409.1556 [cs]}  (\bibinfo{year}{2014}).
  \bibinfo{note}{ArXiv: 1409.1556}.
\bibitem[{Volokitin et~al.(2017)Volokitin, Roig, and
  Poggio}]{volokitin_deep_2017}
\bibinfo{author}{A.~Volokitin}, \bibinfo{author}{G.~Roig},
  \bibinfo{author}{T.~A. Poggio},
\newblock \bibinfo{title}{Do {Deep} {Neural} {Networks} {Suffer} from
  {Crowding}?},
\newblock in: \bibinfo{editor}{I.~Guyon}, \bibinfo{editor}{U.~V. Luxburg},
  \bibinfo{editor}{S.~Bengio}, \bibinfo{editor}{H.~Wallach},
  \bibinfo{editor}{R.~Fergus}, \bibinfo{editor}{S.~Vishwanathan},
  \bibinfo{editor}{R.~Garnett} (Eds.), \bibinfo{booktitle}{Advances in {Neural}
  {Information} {Processing} {Systems} 30}, \bibinfo{publisher}{Curran
  Associates, Inc.}, \bibinfo{year}{2017}, pp. \bibinfo{pages}{5628--5638}.
\bibitem[{LeCun et~al.(1989)LeCun, Boser, Denker, Henderson, Howard, Hubbard,
  and Jackel}]{lecun_backpropagation_1989}
\bibinfo{author}{Y.~LeCun}, \bibinfo{author}{B.~Boser}, \bibinfo{author}{J.~S.
  Denker}, \bibinfo{author}{D.~Henderson}, \bibinfo{author}{R.~E. Howard},
  \bibinfo{author}{W.~Hubbard}, \bibinfo{author}{L.~D. Jackel},
\newblock \bibinfo{title}{Backpropagation {Applied} to {Handwritten} {Zip}
  {Code} {Recognition}},
\newblock \bibinfo{journal}{Neural Computation} \bibinfo{volume}{1}
  (\bibinfo{year}{1989}) \bibinfo{pages}{541--551}.
\bibitem[{Szegedy et~al.(2015)Szegedy, {Wei Liu}, {Yangqing Jia}, Sermanet,
  Reed, Anguelov, Erhan, Vanhoucke, and Rabinovich}]{szegedy_going_2015}
\bibinfo{author}{C.~Szegedy}, \bibinfo{author}{{Wei Liu}},
  \bibinfo{author}{{Yangqing Jia}}, \bibinfo{author}{P.~Sermanet},
  \bibinfo{author}{S.~Reed}, \bibinfo{author}{D.~Anguelov},
  \bibinfo{author}{D.~Erhan}, \bibinfo{author}{V.~Vanhoucke},
  \bibinfo{author}{A.~Rabinovich},
\newblock \bibinfo{title}{Going deeper with convolutions},
\newblock in: \bibinfo{booktitle}{2015 {IEEE} {Conference} on {Computer}
  {Vision} and {Pattern} {Recognition} ({CVPR})}, pp. \bibinfo{pages}{1--9}.
\bibitem[{Guo et~al.(2017)Guo, Dong, Li, and Gao}]{guo_simple_2017}
\bibinfo{author}{T.~Guo}, \bibinfo{author}{J.~Dong}, \bibinfo{author}{H.~Li},
  \bibinfo{author}{Y.~Gao},
\newblock \bibinfo{title}{Simple convolutional neural network on image
  classification},
\newblock in: \bibinfo{booktitle}{2017 {IEEE} 2nd {International} {Conference}
  on {Big} {Data} {Analysis} ({ICBDA})}, pp. \bibinfo{pages}{721--724}.
\bibitem[{Indolia et~al.(2018)Indolia, Goswami, Mishra, and
  Asopa}]{indolia_conceptual_2018}
\bibinfo{author}{S.~Indolia}, \bibinfo{author}{A.~K. Goswami},
  \bibinfo{author}{S.~P. Mishra}, \bibinfo{author}{P.~Asopa},
\newblock \bibinfo{title}{Conceptual {Understanding} of {Convolutional}
  {Neural} {Network}- {A} {Deep} {Learning} {Approach}},
\newblock \bibinfo{journal}{Procedia Computer Science} \bibinfo{volume}{132}
  (\bibinfo{year}{2018}) \bibinfo{pages}{679--688}.
\bibitem[{Li et~al.(2015)Li, Victor, Xiao, and Chen}]{dllectures}
\bibinfo{author}{W.~Li}, \bibinfo{author}{B.~Victor},
  \bibinfo{author}{L.~Xiao}, \bibinfo{author}{H.~Chen}, \bibinfo{title}{Deep
  learning: An overview - lecture notes},
  \bibinfo{howpublished}{"https://studylib.net/doc/15672646/deep-learning--an-overview-university-of-arizona-1"},
  \bibinfo{year}{2015}. \bibinfo{note}{[Online; accessed 10-Jan-2020]}.
\bibitem[{Nguyen et~al.(2016)Nguyen, Tran, Ngo, Phan, Lumbanraja, Faisal,
  Abapihi, Kubo, and Satou}]{nguyen2016dna}
\bibinfo{author}{N.~G. Nguyen}, \bibinfo{author}{V.~A. Tran},
  \bibinfo{author}{D.~L. Ngo}, \bibinfo{author}{D.~Phan},
  \bibinfo{author}{F.~R. Lumbanraja}, \bibinfo{author}{M.~R. Faisal},
  \bibinfo{author}{B.~Abapihi}, \bibinfo{author}{M.~Kubo},
  \bibinfo{author}{K.~Satou},
\newblock \bibinfo{title}{Dna sequence classification by convolutional neural
  network},
\newblock \bibinfo{journal}{Journal of Biomedical Science and Engineering}
  \bibinfo{volume}{9} (\bibinfo{year}{2016}) \bibinfo{pages}{280}.
\bibitem[{Delakis and Garcia(2008)}]{delakis2008text}
\bibinfo{author}{M.~Delakis}, \bibinfo{author}{C.~Garcia},
\newblock \bibinfo{title}{text detection with convolutional neural networks.},
\newblock in: \bibinfo{booktitle}{VISAPP (2)}, pp. \bibinfo{pages}{290--294}.
\bibitem[{Xu and Su(2015)}]{xu2015robust}
\bibinfo{author}{H.~Xu}, \bibinfo{author}{F.~Su},
\newblock \bibinfo{title}{Robust seed localization and growing with deep
  convolutional features for scene text detection},
\newblock in: \bibinfo{booktitle}{Proceedings of the 5th ACM on International
  Conference on Multimedia Retrieval}, \bibinfo{organization}{ACM}, pp.
  \bibinfo{pages}{387--394}.
\bibitem[{Szegedy et~al.(2017)Szegedy, Ioffe, Vanhoucke, and
  Alemi}]{szegedy_inception-v4_2017}
\bibinfo{author}{C.~Szegedy}, \bibinfo{author}{S.~Ioffe},
  \bibinfo{author}{V.~Vanhoucke}, \bibinfo{author}{A.~A. Alemi},
\newblock \bibinfo{title}{Inception-v4, inception-{ResNet} and the impact of
  residual connections on learning},
\newblock in: \bibinfo{booktitle}{Proceedings of the {Thirty}-{First} {AAAI}
  {Conference} on {Artificial} {Intelligence}}, {AAAI}'17,
  \bibinfo{publisher}{AAAI Press}, \bibinfo{address}{San Francisco, California,
  USA}, \bibinfo{year}{2017}, pp. \bibinfo{pages}{4278--4284}.
\bibitem[{Fawaz et~al.(2019)Fawaz, Lucas, Forestier, Pelletier, Schmidt, Weber,
  Webb, Idoumghar, Muller, and Petitjean}]{fawaz_inceptiontime_2019}
\bibinfo{author}{H.~I. Fawaz}, \bibinfo{author}{B.~Lucas},
  \bibinfo{author}{G.~Forestier}, \bibinfo{author}{C.~Pelletier},
  \bibinfo{author}{D.~F. Schmidt}, \bibinfo{author}{J.~Weber},
  \bibinfo{author}{G.~I. Webb}, \bibinfo{author}{L.~Idoumghar},
  \bibinfo{author}{P.-A. Muller}, \bibinfo{author}{F.~Petitjean},
\newblock \bibinfo{title}{{InceptionTime}: {Finding} {AlexNet} for {Time}
  {Series} {Classification}},
\newblock \bibinfo{journal}{arXiv:1909.04939 [cs, stat]}
  (\bibinfo{year}{2019}). \bibinfo{note}{ArXiv: 1909.04939 version: 2}.
\bibitem[{Lines et~al.(2016)Lines, Taylor, and Bagnall}]{lines_hive-cote_2016}
\bibinfo{author}{J.~Lines}, \bibinfo{author}{S.~Taylor},
  \bibinfo{author}{A.~Bagnall},
\newblock \bibinfo{title}{{HIVE}-{COTE}: {The} {Hierarchical} {Vote}
  {Collective} of {Transformation}-{Based} {Ensembles} for {Time} {Series}
  {Classification}},
\newblock in: \bibinfo{booktitle}{2016 {IEEE} 16th {International} {Conference}
  on {Data} {Mining} ({ICDM})}, pp. \bibinfo{pages}{1041--1046}.
  \bibinfo{note}{ISSN: 2374-8486}.
\bibitem[{Bagnall et~al.(2015)Bagnall, Lines, Hills, and
  Bostrom}]{bagnall_time-series_2015}
\bibinfo{author}{A.~Bagnall}, \bibinfo{author}{J.~Lines},
  \bibinfo{author}{J.~Hills}, \bibinfo{author}{A.~Bostrom},
\newblock \bibinfo{title}{Time-{Series} {Classification} with {COTE}: {The}
  {Collective} of {Transformation}-{Based} {Ensembles}},
\newblock \bibinfo{journal}{IEEE Transactions on Knowledge and Data
  Engineering} \bibinfo{volume}{27} (\bibinfo{year}{2015})
  \bibinfo{pages}{2522--2535}. \bibinfo{note}{Conference Name: IEEE
  Transactions on Knowledge and Data Engineering}.
\bibitem[{Brownlee(2018)}]{brownlee_deep_2018}
\bibinfo{author}{J.~Brownlee}, \bibinfo{title}{Deep {Learning} for {Time}
  {Series} {Forecasting}: {Predict} the {Future} with {MLPs}, {CNNs} and
  {LSTMs} in {Python}}, \bibinfo{publisher}{Machine Learning Mastery},
  \bibinfo{year}{2018}. \bibinfo{note}{Google-Books-ID: o5qnDwAAQBAJ}.
\bibitem[{Janos and Roach(2020)}]{pydata_1d_nodate}
\bibinfo{author}{N.~Janos}, \bibinfo{author}{J.~Roach}, \bibinfo{title}{{1D}
  {Convolutional} {Neural} {Networks} for {Time} {Series} {Modeling} - {Nathan}
  {Ja}}, \bibinfo{year}{2020}. \bibinfo{note}{Library Catalog: SlideShare}.
\bibitem[{Alom et~al.(2019)Alom, Taha, Yakopcic, Westberg, Sidike, Nasrin,
  Hasan, Van~Essen, Awwal, and Asari}]{alom_state---art_2019}
\bibinfo{author}{M.~Z. Alom}, \bibinfo{author}{T.~M. Taha},
  \bibinfo{author}{C.~Yakopcic}, \bibinfo{author}{S.~Westberg},
  \bibinfo{author}{P.~Sidike}, \bibinfo{author}{M.~S. Nasrin},
  \bibinfo{author}{M.~Hasan}, \bibinfo{author}{B.~C. Van~Essen},
  \bibinfo{author}{A.~A.~S. Awwal}, \bibinfo{author}{V.~K. Asari},
\newblock \bibinfo{title}{A {State}-of-the-{Art} {Survey} on {Deep} {Learning}
  {Theory} and {Architectures}},
\newblock \bibinfo{journal}{Electronics} \bibinfo{volume}{8}
  (\bibinfo{year}{2019}) \bibinfo{pages}{292}.
\bibitem[{{Xiong} et~al.(2017){Xiong}, {Stiles}, and {Zhao}}]{1DECGSignals}
\bibinfo{author}{Z.~{Xiong}}, \bibinfo{author}{M.~K. {Stiles}},
  \bibinfo{author}{J.~{Zhao}},
\newblock \bibinfo{title}{Robust ecg signal classification for detection of
  atrial fibrillation using a novel neural network},
\newblock in: \bibinfo{booktitle}{2017 Computing in Cardiology (CinC)}, pp.
  \bibinfo{pages}{1--4}.
\bibitem[{Khan et~al.(2018)Khan, Rahmani, Shah, Bennamoun, Medioni, and
  Dickinson}]{khan_guide_2018}
\bibinfo{author}{S.~Khan}, \bibinfo{author}{H.~Rahmani},
  \bibinfo{author}{S.~A.~A. Shah}, \bibinfo{author}{M.~Bennamoun},
  \bibinfo{author}{G.~Medioni}, \bibinfo{author}{S.~Dickinson},
  \bibinfo{title}{A {Guide} to {Convolutional} {Neural} {Networks} for
  {Computer} {Vision}}, \bibinfo{publisher}{Morgan \& Claypool},
  \bibinfo{year}{2018}.
\bibitem[{Saha(2018)}]{saha_comprehensive_2018}
\bibinfo{author}{S.~Saha}, \bibinfo{title}{A {Comprehensive} {Guide} to
  {Convolutional} {Neural} {Networks} — the {ELI}5 way},
  \bibinfo{year}{2018}.
\bibitem[{{Son Lam Phung} and {Abdesselam
  Bouzerdoum}(2009)}]{son_lam_phung_matlab_2009}
\bibinfo{author}{{Son Lam Phung}}, \bibinfo{author}{{Abdesselam Bouzerdoum}},
  \bibinfo{title}{{MATLAB} {Library} for {Convolutional} {Neural} {Networks}},
  \bibinfo{type}{Technical {Report}}, \bibinfo{address}{Visual and Audio Signal
  Processing Lab, University of Wollongong}, \bibinfo{year}{2009}.
\bibitem[{Stutz(2014)}]{stutz_understanding_2014}
\bibinfo{author}{D.~Stutz}, \bibinfo{title}{Understanding {Convolutional}
  {Neural} {Networks}}, \bibinfo{type}{Seminar {Report}}, \bibinfo{year}{2014}.
\bibitem[{Lichman(2013)}]{lichman_uci_2013}
\bibinfo{author}{M.~Lichman}, \bibinfo{title}{{UCI} {Machine} {Learning}
  {Repository}}, \bibinfo{publisher}{University of California, Irvine, School
  of Information and Computer Sciences}, \bibinfo{year}{2013}.
\bibitem[{{{CNN - Matlab}}(2019)}]{noauthor_convolutional_nodate}
\bibinfo{author}{{{CNN - Matlab}}}, \bibinfo{title}{Convolutional {Neural}
  {Network}}, \bibinfo{howpublished}{retrieved from,
  \url{https://au.mathworks.com/solutions/deep-learning/convolutional-neural-network.html}},
  \bibinfo{year}{2019}.
\bibitem[{Brownlee(2019)}]{brownlee_gentle_2019}
\bibinfo{author}{J.~Brownlee}, \bibinfo{title}{A {Gentle} {Introduction} to
  {Pooling} {Layers} for {Convolutional} {Neural} {Networks}},
  \bibinfo{howpublished}{retrieved from,
  \url{https://machinelearningmastery.com}}, \bibinfo{year}{2019}.
\bibitem[{{{Convolutional {Neural} {Networks} ({LeNet}) — {DeepLearning} 0.1
  documentation - }{CNN - LeNet}}(2019)}]{noauthor_convolutional_nodate-2}
\bibinfo{author}{{{Convolutional {Neural} {Networks} ({LeNet}) —
  {DeepLearning} 0.1 documentation - }{CNN - LeNet}}},
  \bibinfo{title}{Convolutional {Neural} {Networks} ({LeNet})},
  \bibinfo{howpublished}{retrieved from
  \url{http://deeplearning.net/tutorial/lenet.html}}, \bibinfo{year}{2019}.
\bibitem[{Almufti(2019)}]{almufti_historical_2019}
\bibinfo{author}{S.~M. Almufti},
\newblock \bibinfo{title}{Historical survey on metaheuristics algorithms},
\newblock \bibinfo{journal}{International Journal of Scientific World}
  \bibinfo{volume}{7} (\bibinfo{year}{2019}) \bibinfo{pages}{1--12}.
\bibitem[{Goldberg(1989)}]{goldberg_genetic_1989}
\bibinfo{author}{D.~E. Goldberg}, \bibinfo{title}{Genetic algorithms in search,
  optimization, and machine learning}, \bibinfo{publisher}{Addison-Wesley
  Longman Publishing Co., Inc.}, \bibinfo{address}{Boston, MA, USA},
  \bibinfo{edition}{1st} edition, \bibinfo{year}{1989}.
\bibitem[{Eberhart and Kennedy(1995)}]{eberhart_new_1995}
\bibinfo{author}{R.~Eberhart}, \bibinfo{author}{J.~Kennedy},
\newblock \bibinfo{title}{A new optimizer using particle swarm theory},
\newblock in: \bibinfo{booktitle}{Micro {Machine} and {Human} {Science}, 1995.
  {MHS} '95., {Proceedings} of the {Sixth} {International} {Symposium} on}, pp.
  \bibinfo{pages}{39 --43}.
\bibitem[{Sharma(2010)}]{sharma_new_2010}
\bibinfo{author}{A.~Sharma},
\newblock \bibinfo{title}{A new optimizing algorithm using reincarnation
  concept},
\newblock in: \bibinfo{booktitle}{11th {IEEE} {International} {Symposium} on
  {Computational} {Intelligence} and {Informatics} ({CINTI})}, pp.
  \bibinfo{pages}{281 --288}.
\bibitem[{Bhardwaj and Tiwari(2015)}]{bhardwaj_breast_2015}
\bibinfo{author}{A.~Bhardwaj}, \bibinfo{author}{A.~Tiwari},
\newblock \bibinfo{title}{Breast cancer diagnosis using {Genetically}
  {Optimized} {Neural} {Network} model},
\newblock \bibinfo{journal}{Expert Systems with Applications}
  \bibinfo{volume}{42} (\bibinfo{year}{2015}) \bibinfo{pages}{4611--4620}.
\bibitem[{Chen et~al.(2011)Chen, Yang, Liu, and Liu}]{chen_support_2011}
\bibinfo{author}{H.-L. Chen}, \bibinfo{author}{B.~Yang},
  \bibinfo{author}{J.~Liu}, \bibinfo{author}{D.-Y. Liu},
\newblock \bibinfo{title}{A support vector machine classifier with rough
  set-based feature selection for breast cancer diagnosis},
\newblock \bibinfo{journal}{Expert Systems with Applications}
  \bibinfo{volume}{38} (\bibinfo{year}{2011}) \bibinfo{pages}{9014--9022}.
\bibitem[{Zheng et~al.(2014)Zheng, Yoon, and Lam}]{zheng_breast_2014}
\bibinfo{author}{B.~Zheng}, \bibinfo{author}{S.~W. Yoon},
  \bibinfo{author}{S.~S. Lam},
\newblock \bibinfo{title}{Breast cancer diagnosis based on feature extraction
  using a hybrid of {K}-means and support vector machine algorithms},
\newblock \bibinfo{journal}{Expert Systems with Applications}
  \bibinfo{volume}{41} (\bibinfo{year}{2014}) \bibinfo{pages}{1476--1482}.
\bibitem[{Liu et~al.(2009)Liu, Wang, and Zhang}]{liu_decision_2009}
\bibinfo{author}{Y.-Q. Liu}, \bibinfo{author}{C.~Wang},
  \bibinfo{author}{L.~Zhang},
\newblock \bibinfo{title}{Decision {Tree} {Based} {Predictive} {Models} for
  {Breast} {Cancer} {Survivability} on {Imbalanced} {Data}},
\newblock in: \bibinfo{booktitle}{2009 3rd {International} {Conference} on
  {Bioinformatics} and {Biomedical} {Engineering}}, pp. \bibinfo{pages}{1--4}.
  \bibinfo{note}{ISSN: 2151-7622}.
\bibitem[{Karabatak(2015)}]{karabatak_new_2015}
\bibinfo{author}{M.~Karabatak},
\newblock \bibinfo{title}{A new classifier for breast cancer detection based on
  {Naïve} {Bayesian}},
\newblock \bibinfo{journal}{Measurement} \bibinfo{volume}{72}
  (\bibinfo{year}{2015}) \bibinfo{pages}{32--36}.
\bibitem[{Arute et~al.(2019)Arute, Arya, Babbush, Bacon, Bardin, Barends, and
  Biswas}]{arute_quantum_2019}
\bibinfo{author}{F.~Arute}, \bibinfo{author}{K.~Arya},
  \bibinfo{author}{R.~Babbush}, \bibinfo{author}{D.~Bacon},
  \bibinfo{author}{J.~C. Bardin}, \bibinfo{author}{R.~Barends},
  \bibinfo{author}{J.~M. Biswas},
\newblock \bibinfo{title}{Quantum supremacy using a programmable
  superconducting processor},
\newblock \bibinfo{journal}{Nature} \bibinfo{volume}{574}
  (\bibinfo{year}{2019}) \bibinfo{pages}{505--510}.

\end{thebibliography}







\end{document}